%% file: paper-main.tex
\documentclass[lettersize,journal]{IEEEtran}

\usepackage[utf8]{inputenc} %
\usepackage[T1]{fontenc}    %
\usepackage[hypertexnames=false]{hyperref} %
\usepackage{url}            %
\usepackage{booktabs}       %
\usepackage{array}          %
\usepackage{amsfonts}       %
\usepackage{nicefrac}       %
\usepackage{microtype}      %
\usepackage{graphicx}       %
\graphicspath{{media/}}     %
\usepackage{amsmath}        %
\usepackage{xcolor}         %
\usepackage{listings}       %
\usepackage{algorithm}      %
\usepackage{algpseudocode}  %
\usepackage{pdflscape}      %
\usepackage[caption=false,font=footnotesize]{subfig} %
\usepackage{stfloats}       %
\usepackage{circuitikz}     %
\usetikzlibrary{arrows.meta}
\usepackage{standalone}     %

\usepackage{orcidlink}

\lstdefinestyle{db}{
  basicstyle=\ttfamily\footnotesize,
  columns=fullflexible,
  frame=single,
  breaklines=true,
  showstringspaces=false,
  keepspaces=true,
  captionpos=b,
  aboveskip=3pt plus 1pt,
  belowskip=3pt plus 1pt,
}
\definecolor{dslkey}{RGB}{25,80,150}
\definecolor{dslval}{RGB}{25,25,25}
\definecolor{dslstr}{RGB}{140,60,20}
\definecolor{dslcmt}{RGB}{115,115,115}
\newcommand\YAMLkeystyle{\ttfamily\footnotesize\color{dslkey}}
\newcommand\YAMLvaluestyle{\ttfamily\footnotesize\color{dslval}}
\newcommand\YAMLcolonstyle{\color{dslval}\mdseries}
\lstdefinelanguage{yaml}{
  keywords={true,false,null},
  sensitive=false,
  basicstyle=\YAMLkeystyle,
  comment=[l]{\#},
  commentstyle=\color{dslcmt}\itshape,
  morestring=[b]',
  morestring=[b]",
  stringstyle=\color{dslstr},
  moredelim=**[il][\YAMLcolonstyle{:}\YAMLvaluestyle]{:},
  literate={>}{{\textcolor{dslstr}{\textgreater}}}1
           {-\ }{{\textcolor{dslval}{-\ }}}2,
}
\lstdefinelanguage{json}{
  sensitive=false,
  comment=[l]{//},
  commentstyle=\color{dslcmt}\itshape,
  morestring=[b]",
  stringstyle=\color{dslkey},
  literate={:}{{{\color{dslval}:}}}1
           {,}{{{\color{dslval},}}}1,
}

\title{Analog-DB: An Agent-First Analog Integrated Circuit Database, From Blocks to Systems}

\author{Danial~Noori~Zadeh~\orcidlink{0009-0001-8998-3343} and Mohamed~B.~Elamien~\orcidlink{0000-0002-6528-0010}%
\thanks{The authors are with the Department of Electrical and Computer
Engineering, McMaster University, Hamilton, ON, Canada
(e-mail: noorizad@mcmaster.ca, dnoorizadeh@gmail.com;
elamienm@mcmaster.ca).}}

\begin{document}
\maketitle

\begin{abstract}
Sharing analog integrated circuit designs remains difficult: foundry non-disclosure agreements restrict the process details a design depends on, and the testbenches behind published results are rarely released. We present \emph{analog-db}, an open-source, versioned database built on a shareable design representation. A domain-specific language captures each design as a process-neutral topology, reusable testbenches, and a machine-readable datasheet under one schema, so a design is shared in full and re-simulates on the process kits it is bound to. A parameterization scheme exposes functional sub-blocks and device sizes as named parameters that carry their matching constraints, making circuits composable and retargetable; a schema-governed contract and queryable catalog let AI design agents discover and reuse them directly. Across the regulator corpus, all 23 circuit-kit bindings on three open kits meet their own recorded specification bands (typical corner, matched devices, no layout) and 10 of 23 meet a common class band. Seventeen of the 23 imported sizings failed their testbenches and closed under a \(g_m/I_D\) sizing loop driven by the annotated sub-block roles, typically within one to three iterations. In a supervised case study, a coding agent working from the released artifacts sized the op-amp cores of a chopper instrumentation amplifier on an open 130\,nm kit, locating four hand-entry defects and a missing common-mode feedback loop that the sizing-only baseline did not repair. The database holds 68 circuits across sixteen classes, verifiable at schematic level under a tiered harness and tracked on a power/performance scoreboard, released at \url{https://github.com/MacAnalog/spicexplorer-release}.
\end{abstract}

\begin{IEEEkeywords}
Electronic Design Automation, Analog IC Design, Open-Source PDK, Design Reuse, LLM Agents, Capacitively Coupled Instrumentation Amplifier, Low Dropout Regulator, Biomedical Frontend, Design Benchmarks.
\end{IEEEkeywords}

\section{Introduction}
\label{sec:intro}

\IEEEPARstart{A}{nalog} integrated circuits are essential to systems as far apart as high-bandwidth
memory interfaces~\cite{general_hbm_example_chae4nm115TB2024a}, radio-frequency
transceivers, and low-power
biomedical interfaces, from neural-recording front
ends~\cite{general_biomedical_example_harrison2003low} to
wearables~\cite{general_biomedical_review_koo2025design}. Yet industry
and academia have automated digital blocks far more successfully than analog
ones. Part of the difficulty is that an analog block's performance rests on
tradeoffs among gain, bandwidth, dynamic range, noise, power, and
area~\cite{general_sansen2006analog, general_danial_genAI_review, general_tradeoff_wangScalingOptimumDynamic2019}: no circuit is best
on all of them at once, only best for a target application. A designer explores these manually and
reports the final numbers. Analog design, unlike digital, still has no common
way to share these circuits together with the testbenches that validate them;
two structural barriers explain why.

The first barrier is representational: a netlist carries connectivity, not
intent. Subcircuits nest to any depth, so hierarchy itself is expressible; what
no released artifact states is which devices form a cascode current mirror
rather than a simple one, which stage sets the gain, or which path closes a
feedback loop. A fully differential OTA may combine several stages, each with
its own compensation architecture, common-mode feedback network, and bias
network, and space limits keep most papers from publishing the full working
circuit~\cite{analogdb_template_CMFB_banuFullyDifferentialOperational1988}, let
alone the reasoning that names its parts. The omission compounds up the
hierarchy: once that OTA is embedded in an instrumentation amplifier, a
low-dropout regulator, or an analog-to-digital converter, system-level papers
treat it as a solved sub-block. Fan et al., for instance, report a capacitively
coupled chopper instrumentation amplifier (CCIA) for wireless sensor nodes and
publish block- and simplified transistor-level schematics, but not the device
sizes, bias values, or testbenches behind the reported
numbers~\cite{analogdb_design_chopper_ccia_fan18$mu$W602011}. A later
designer can cite that amplifier but cannot rebuild it.

The same gap afflicts verification. A digital block travels with checks
that are themselves readable: assertions and expected logic waveforms state
a pass condition. Verifying an analog block instead requires measurement
recipes over continuous waveforms, in which the stimulus, the corner, the
extraction, and the threshold separating pass from fail are each a
separate, unrecorded decision. The reader thus receives the achieved
performance figures, but not the procedure that measured them.

The second barrier is contractual: the artifacts that would make a published
result checkable are the ones an author may not release. Foundry
non-disclosure agreements encumber the process models a design depends on, so
authors can publish neither the exact netlist nor the simulation results
behind a reported number, although recent work proposes frameworks for
secure redaction of such NDA-encumbered
information~\cite{intro_nda_liSABLENDASafeClosedLoop2026}.

Both barriers are answerable now in a way they were not a decade ago. Open
process design kits (PDKs) have made a growing class of designs freely
simulable (\S\ref{sec:related}): a designer who releases against such a kit
hands the reader a circuit they can run. This removes the contractual
barrier but not the representational one, since nothing in a process kit
records how a published circuit is organized or under what testbench its
numbers were obtained. The field gained a process to share against; it
still lacks a place to put the designs.

A second opportunity comes from language-model agents, now applied to
analog sizing, topology selection, and subcircuit
identification~\cite{agentic_AnalogSAGE,EDA_sizer_LLM_BO_HEART,agentic_phamGENIEASIGenerativeInstruction2025}.
Such an agent needs an engineered context: a hierarchical design together
with the testbench that exercises it and the recorded result that says
what it achieved. The three together let an agent reason about
topologies and judge whether a candidate improves on what already exists,
and no released corpus we are aware of supplies all three in one artifact
(\S\ref{sec:related}). This demand motivates the
machine-actionability requirement of \S\ref{sec:requirements}.

\begin{figure*}[t]
  \centering
  \includegraphics[width=\linewidth]{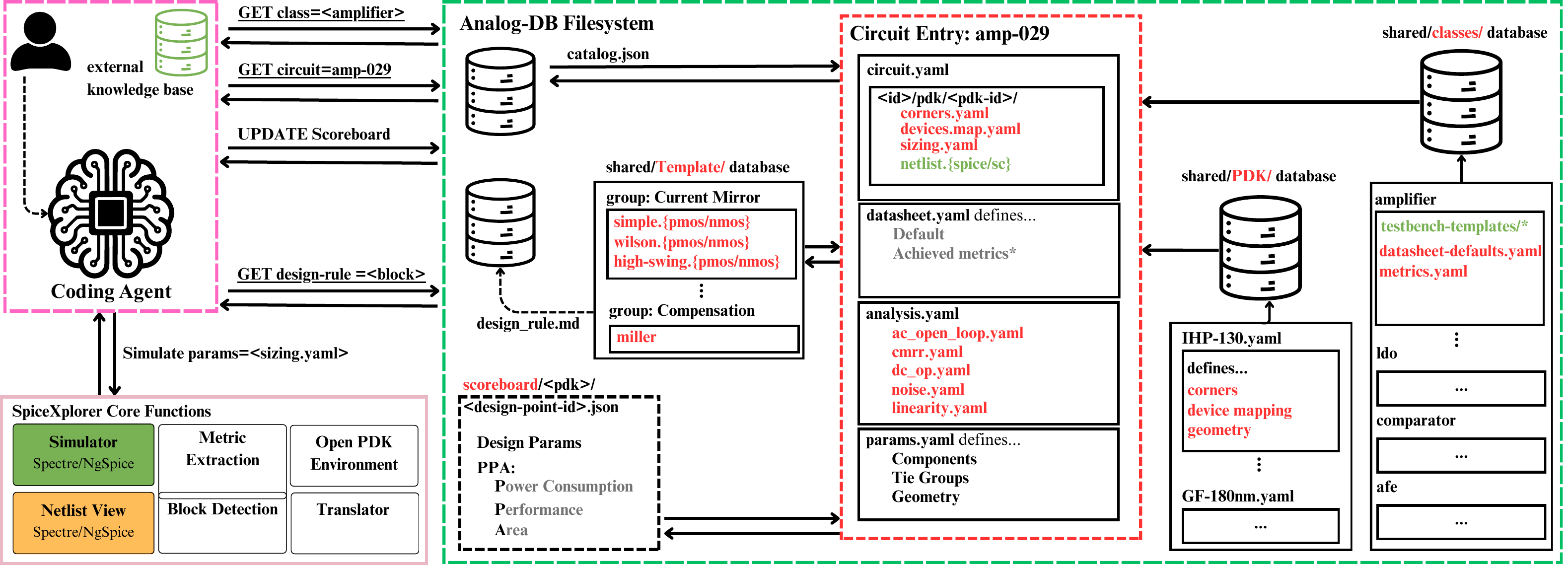}
  \caption{System overview. A design agent queries the catalog by class
    and accession identifier, retrieves an entry (topology, datasheet,
    analyses, parameters), and simulates it through the SpiceXplorer
    harness; entries draw on shared class definitions and process-kit
    registries. File labels in the drawing are illustrative rather than
    verbatim: the released kit registries are \texttt{ihp-sg13g2},
    \texttt{gf180mcu}, and \texttt{sky130}; per-kit device geometry
    is stored in each entry's \texttt{sizing.yaml}; and every released
    simulation runs on ngspice against the open kits (the other
    simulator lane shown ships kit-unbound).}
    \label{fig:overview}
\end{figure*}

One representation addresses both barriers: it separates what
is shareable from what is encumbered and names the structure and acceptance
conditions a netlist leaves implicit. We present
\emph{analog-db}\footnote{Written \emph{analog-db} throughout, after the
released artifact's name; the title's \emph{Analog-DB} is the same name
capitalized.}, an
open-source, versioned database built on such a representation. A design
can then be released in full: a reader who holds the relevant kit can
re-simulate it directly, and one who does not can port it to a kit they
own. Because the representation names a circuit's sub-blocks rather than
flattening them, that design can also be reused inside a larger one rather
than re-derived. Our contributions are as follows:

\begin{itemize}
  \item \textbf{A foundry-neutral sharing method.} A domain-specific language that
        captures each design as a PDK-neutral topology, a reusable testbench, and a
        machine-readable datasheet under one schema and carries no proprietary process
        detail, so a design can be released in full, re-simulated directly on any
        kit it binds, and retargeted to another by adding one binding.
  \item \textbf{An agent-first, human-readable interface.} A schema-governed contract
        and a queryable catalog addressable by AI design agents without a service,
        paired with auto-generated schematic views of the same entries.
  \item \textbf{A portability and verification path.} Automatic lowering to multiple
        open PDKs as verified SPICE decks, cross-PDK retargeting, a
        tiered verification harness, and a power/performance scoreboard
        (with the gate-area proxy of \S\ref{sec:scoreboard}), so a
        released design carries a checkable status and can be retargeted to the
        process the reader has.
  \item \textbf{A sub-block annotation and parameterization scheme.} An annotation of
        each design's detected functional sub-blocks (current mirrors, differential
        pairs, cross-coupled loads; 51\% of corpus devices at the archived
        revision) alongside a complete parameterization that exposes device sizes as named parameters
        carrying their matching constraints, making circuits composable and
        retargetable.
  \item \textbf{An open, populated database.} 68 verifiable circuits across sixteen
        classes, 65 of them carrying recorded design points (schematic level,
        typical corner, matched devices) (Table~\ref{tab:coverage}), released as a versioned corpus that
        others can extend.
\end{itemize}

\section{Background and Related Work}
\label{sec:related}
Existing efforts each supply part of what reuse demands, and none
supplies all of it in a released corpus. Generator frameworks bind a
circuit to its own testbench and measurement managers and carry them
across processes, but they distribute generators rather than a curated,
queryable corpus with recorded acceptance
criteria~\cite{EDA_generator_bag2_changBAG2ProcessPortable2018}.
PDK-neutral intermediate representations give the topology a schema and
put the analyses in the same object model, across the same open kits
used here, without shipping designs, results, or structural
annotation~\cite{EDA_ir_vlsir_fritchmanVLSIRModularFramework2023}.
Structural annotation of flat netlists is an established line of its
own, by exact subgraph isomorphism, by learned models, and by
library-free bottom-up construction, but it operates on netlists the
user supplies rather than on a released corpus that carries its
annotations as part of the
record~\cite{EDA_annotation_gana_kunalGANAGraphConvolutional2020,EDA_annotation_gnn_kunalGNNBasedHierarchical2023,EDA_annotation_libfree_neunerLibraryFreeStructure2021}.
Benchmark corpora release circuits at far larger scale than this work,
one with recorded simulated
results~\cite{EDA_benchmark_cktgnn_dongCktGNNCircuitGraph2023}, the
other drawn from the literature without
testbenches~\cite{EDA_benchmark_amsnet2_shiAMSnetLargeAMS2025}, and
testing suites release circuits with shared testbenches on an open
kit~\cite{analogdb_design_AnalogGym_liAnalogGymOpenPractical2024}, but
none of them records per-circuit acceptance bands, hierarchy, or a
machine-readable contract. The contribution of analog-db is the
integration: one versioned corpus in which the topology, the testbench,
the acceptance band, the recorded outcome, and the structural
annotation are the same object, addressable by an agent without a
service. Table~\ref{tab:priorwork} compares the closest efforts.

\begin{table*}[t]
  \caption{analog-db against the closest released artifacts. Each cell
  states what that system ships as released, taken from its publication
  and public repository; ``--'' marks a capability it does not
  provide.}\label{tab:priorwork}
  \centering
  \footnotesize
  \setlength{\tabcolsep}{4pt}
  \begin{tabular}{@{}p{2.2cm} p{3.0cm} p{2.8cm} p{2.9cm} p{2.8cm} p{2.5cm}@{}}
    \toprule
    System & Unit of release, scale & Testbenches shipped & Recorded results, acceptance bands & Structure, hierarchy & Machine contract \\
    \midrule
    AnalogGym~\cite{analogdb_design_AnalogGym_liAnalogGymOpenPractical2024} &
    Testing suite; 30 topologies, 5 categories; 2 of 5 categories run
    open-source on the bundled sky130 kit &
    Yes: 2 shared templates, amplifier and LDO only &
    Sizing-method baselines; no per-circuit acceptance bands &
    Flat netlists; no composition &
    -- \\
    CORA-OpAmp~\cite{example_cora_opamp} &
    RL sizing flow; 1 topology (folded cascode) on an open kit &
    Yes, its one topology &
    Single design point &
    -- &
    -- \\
    OCB~\cite{EDA_benchmark_cktgnn_dongCktGNNCircuitGraph2023} &
    Topology dataset; 10{,}000 behavioral op-amps &
    Generation and evaluation harness; no per-circuit testbench &
    Simulated results; no acceptance bands &
    Graph encoding of each topology &
    -- \\
    AMSnet~2.0~\cite{EDA_benchmark_amsnet2_shiAMSnetLargeAMS2025} &
    Schematic/netlist database; 2{,}686 circuits from the literature &
    -- &
    -- &
    Schematic-netlist correspondence &
    -- \\
    AMS-SizingBench~\cite{EDA_benchmark_amssizingbench_yuAutoSizerAutomaticSizing2026} &
    Sizing benchmark; 24 AMS circuits on SKY130 &
    Simulator-based evaluation under stated constraints &
    Per-circuit spec thresholds; no recorded design points &
    -- &
    -- \\
    Masala-CHAI~\cite{EDA_benchmark_masalachai_bhandariMasalaCHAILargeScale2024} &
    Netlist dataset; 7{,}500 schematics from 10 textbooks &
    -- &
    Netlist-level verification; no recorded performance &
    Schematic-netlist correspondence &
    -- \\
    AnalogGenie~\cite{EDA_benchmark_analoggenie_gaoAnalogGenieGenerative2025} &
    Generative topology engine with its training dataset &
    -- &
    -- &
    Sequence-based graph representation &
    -- \\
    BAG2~\cite{EDA_generator_bag2_changBAG2ProcessPortable2018} &
    Generator framework; no fixed corpus &
    Generator-embedded testbench and measurement managers &
    Re-measured per instantiation; no released corpus of outcomes &
    Hierarchical generators &
    Programmatic API \\
    VLSIR~\cite{EDA_ir_vlsir_fritchmanVLSIRModularFramework2023} &
    PDK-neutral IR with published open-kit packages &
    Analyses in the same object model &
    -- &
    Schema-governed hierarchy &
    Protobuf schema \\
    \midrule
    analog-db (this work) &
    Versioned corpus; 68 circuits, 16 classes, 3 open kits &
    Yes: 65 of 68 circuits bind class-owned templates; 3 fall outside
    them\(^{a}\) &
    Recorded design points judged against per-metric specification
    bands &
    Detected sub-blocks; composition closed over entries &
    Schema-governed file contract + catalog \\
    \bottomrule
  \end{tabular}
  \vspace{2pt}

  {\footnotesize \(^{a}\)The ideal-amplifier programmable-gain stage
  (\texttt{ia\_005}) and the binary capacitor bank (\texttt{sw\_003})
  have static code pins outside the class port contract, so
  their measured values are recorded in the datasheet without a bound
  bench (Table~\ref{tab:ia-ihp}); the line driver (\texttt{drv\_001})
  binds self-contained testbench decks through the optimizer
  projection instead of class templates.\par}
\end{table*}

AnalogGym~\cite{analogdb_design_AnalogGym_liAnalogGymOpenPractical2024}
is the closest testing suite. It ships 30 topologies in five categories,
two shared testbench templates that cover its amplifier and
low-dropout-regulator categories, one uniform extraction script over AC,
DC, and transient measures, and a bundled sky130 kit on which two of
the five, the amplifiers and regulators, run open-source as
distributed; the sensing-front-end, voltage-reference, and
phase-locked-loop categories ship no open-source-runnable decks. What
it does not record is what reuse needs
next: no noise, distortion, or intermodulation analyses, no per-circuit
acceptance bands, no way for one block to compose into a larger one, and
no machine-readable contract. CORA-OpAmp~\cite{example_cora_opamp} shows
that a reinforcement-learning sizing loop can be made reproducible on an
open PDK, but at a narrow scope: it targets a single folded-cascode
topology and drives the simulator through scripting written for that
topology.
Three recent dataset efforts push scale on complementary axes, and each
stops short of a different part of what reuse needs.
AMS-SizingBench~\cite{EDA_benchmark_amssizingbench_yuAutoSizerAutomaticSizing2026}
releases 24 mixed-signal circuits on SKY130 with per-circuit
specification thresholds and a simulator-in-the-loop harness, but the
thresholds ship without recorded design points that meet them, and no
entry composes into a larger one.
Masala-CHAI~\cite{EDA_benchmark_masalachai_bhandariMasalaCHAILargeScale2024}
extracts 7{,}500 textbook schematics into verified SPICE netlists, two
orders of magnitude beyond the corpus released here, but its unit of
release is the
bare netlist: it ships without testbenches, recorded results, or a
machine contract.
AnalogGenie~\cite{EDA_benchmark_analoggenie_gaoAnalogGenieGenerative2025}
generates circuit topologies over a sequence-based graph dataset, and
the generated artifact is again connectivity alone rather than a
verified, benched entry.
Teaching corpora such as Pretl's analog circuit design
collection~\cite{general_Pretl_AICD_GitHub} publish worked designs on the
same open tools, but as course material rather than as a verified,
queryable corpus. Adjacent open-source flows automate complementary
stages of the same pipeline: OpenFASoC generates mixed-signal blocks end
to end~\cite{EDA_flow_openfasoc_github}, ALIGN automates analog
layout~\cite{EDA_layout_align_kunalALIGNOpenSource2019}, AnalogCoder
generates topologies with language
models~\cite{EDA_agent_analogcoder_laiAnalogCoderAnalog2025}, and
AICircuit benchmarks learned sizing across circuit
levels~\cite{EDA_benchmark_aicircuit_mehradfarAICircuitMultiLevel2024}.

Automated analog design predates language-model agents by decades. A long
line of work drives sizing with evolutionary and Bayesian search, much of
its methodological effort spent shaping cost functions and handling
constraints across competing
metrics~\cite{score_adaptive_normalized,score_gaussian_fitness_function,score_LoCoMOBO_,score_efficient_implicit_constraint_handling,score_EAVREF_product_FoM},
and symbolic analysis explores topology choices
analytically~\cite{general_danial_symxplorer}. Cross-node re-sizing is
likewise an established subfield: constant-inversion-level resizing
rules~\cite{EDA_reuse_resizing_galupmontoroResizingRulesMOS2002} are in
substance the \(g_m/I_D\)-preserving retarget this work applies per
detected role (\S\ref{sec:lowering}). Language models extend the
annotation line, identifying the functional sub-blocks of a flat
netlist~\cite{agentic_phamGENIEASIGenerativeInstruction2025}, a problem
analog-db's deterministic template matching addresses at release time
(\S\ref{sec:topology}). Agent-driven sizing flows, meanwhile, already
presuppose a corpus of this kind.
AnalogSAGE~\cite{agentic_AnalogSAGE} and HeaRT, a hierarchical
circuit-reasoning agentic framework that couples reasoning-tree agents
with off-the-shelf optimizers, evaluated on its authors' 40-circuit
benchmark of flattened SPICE
netlists~\cite{EDA_sizer_LLM_BO_HEART}, both drive a simulator from an agent
loop, yet neither releases the designs or the testbenches it ran on. Each
therefore assembles a private corpus that does not outlive the paper
reporting it: a later agent can read the numbers but cannot re-simulate
the flow that produced them. analog-db answers this demand by exposing
every entry through a schema-governed set of released files
(\S\ref{sec:agent}) that an agent can query directly.

Open process design kits changed what a published analog result can carry
with it: a design released against one is simulable by any reader who
downloads the kit, with no foundry entitlement. An open kit, however, fixes only the model a netlist is
evaluated against, not the netlist's own legibility.
The representation of \S\ref{sec:representation} supplies that missing
layer and lowers to the open kits, so its entries can be simulated
with them.

\section{Method}
\label{sec:method}

\subsection{Requirements for a Shareable Design Representation}
\label{sec:requirements}
The two barriers of Section~\ref{sec:intro} translate directly into the first
two requirements below. The remaining three follow from the opportunity that
makes a shared corpus worth building, namely reuse across the design hierarchy
and reuse by software agents. Together they define what a shareable analog
design representation must satisfy, each addressed by the subsection noted:
\begin{itemize}
  \item \textbf{R1: Foundry neutrality.} A released design carries no
        foundry-encumbered process detail, yet remains simulable through open
        process kits (\S\ref{sec:topology}).
  \item \textbf{R2: Reproducibility.} The testbenches, extraction recipes, and
        acceptance criteria behind every reported result travel with the design
        itself (\S\ref{sec:datasheet}).
  \item \textbf{R3: Structure over opacity.} A design exposes its functional
        sub-blocks and its named parameters with their matching constraints,
        rather than an opaque flat
        netlist (\S\ref{sec:topology}, \S\ref{sec:params}).
  \item \textbf{R4: Composability.} Verified blocks compose into larger systems that
        are again verifiable entries of the same kind
        (\S\ref{sec:composition}).
  \item \textbf{R5: Machine actionability.} Every artifact is schema-governed and
        queryable, so a software agent can discover, compare, and reuse designs
        with a verifiable status, while the same artifacts stay readable to
        human designers (\S\ref{sec:agent}).
\end{itemize}
Section~\ref{sec:portability}
then describes how generation and verification keep R1 and R2 checkable for
every released artifact.

\subsection{The analog-db Design Representation}
\label{sec:representation}

\subsubsection{A Formal Model of a Design Entry}
\label{sec:overview}
The unit of release in analog-db is the \emph{design entry},
\begin{equation}
  E = (T,\, \Theta,\, C,\, B),
  \label{eq:entry}
\end{equation}
a four-part object: a process-neutral \emph{topology} $T$, a
\emph{parameter space} $\Theta$ carrying the matching constraints among
devices (\S\ref{sec:params}), a \emph{behavioral contract} $C$, and a set
of per-process \emph{bindings} $B$. The first three are detailed in the
subsections that follow. A \emph{binding}
$b \in B$
supplies everything one process kit requires: a mapping from generic device
kinds to kit devices, concrete sizing values with search bounds, and corner
selections. \emph{Lowering} is then a mechanical function
\begin{equation}
  \mathcal{L}\colon (T, a, b) \longmapsto \mathrm{deck}_{a,b},
  \label{eq:lowering}
\end{equation}
where $a$ ranges over the analyses the contract declares and $b$ over the
entry's bindings, so the same entry yields one ready-to-run SPICE deck per
bound kit and analysis.

Three further terms recur throughout the paper. A \emph{design point} is a
recorded simulation outcome: one concrete sizing of one entry on one kit,
stored with its per-corner measured metrics and pass/fail verdicts under a
content-derived identifier, so every reported number remains reproducible from
its record. A \emph{validated design point} is a
design point whose every spec-bounded metric passes, the T4 criterion of
\S\ref{sec:verify}; the paper uses both terms in exactly these
senses. An \emph{accession identifier} is the stable, append-only name
under which an entry is published and cited. Throughout, artifacts that encode
designer intent are authored and human-reviewed, while derived artifacts are
regenerated mechanically and checked against drift
(\S\ref{sec:verify}). Figure~\ref{fig:overview} shows how an agent interacts
with these pieces, and Appendix~\ref{app:dsl} gives the concrete syntax of
each artifact on a worked example.

\subsubsection{Separating Topology from Process}
\label{sec:topology}
\begin{algorithm}[t]
  \caption{Sub-block detection by labeled subgraph matching.}\label{alg:block-detection}
  \begin{algorithmic}[1]
    \Require flattened design graph $G$; template library $\mathcal{T}$
      (Table~\ref{tab:templates})
    \Ensure block instances $R$; per-device structural roles
    \State label every graph: nodes by device kind and polarity; edges by
      their gate/drain/source pin multiset, so a diode connection forms
      parallel edges; bulk pins only where a template demands them
    \State $R \gets \emptyset$
    \ForAll{templates $t \in \mathcal{T}$ with graph $G_t$}
      \ForAll{label-preserving monomorphisms $\varphi\colon G_t \hookrightarrow G$}
        \If{supply rails anchor as $t$ demands \textbf{and} each internal
          net of $t$ maps to a private net of equal degree}
          \State add block instance $(t, \varphi)$ to $R$
        \EndIf
      \EndFor
    \EndFor
    \State on an equal device set, prefer a bulk-aware match to a bulk-blind one
    \State assign each covered device the structural role its match names
    \State render the annotated and hierarchical views from $R$
  \end{algorithmic}
\end{algorithm}
A circuit's topology can be stated without any process detail, and separating
the two is what makes a design releasable (R1). In the abstract netlist,
every device is a generic token carrying only its kind and polarity, and
every geometry field refers to a named symbol rather than a number; nothing
identifies a foundry device, model section, or corner library. All such
detail enters through the per-kit binding, so the released topology is free
of foundry-encumbered content. A reader who holds any
bound kit obtains a simulable circuit by lowering, while a reader who holds
none can still read, cite, and retarget the design.

A flat netlist, however, satisfies R1 without satisfying R3. The
representation therefore recovers the missing structure mechanically, by
labeled subgraph matching of a functional template library against the
flattened design graph (Algorithm~\ref{alg:block-detection}). Each match
assigns the covered devices a
deterministic structural role, and the assigned roles drive two generated
views of the same entry: an annotated schematic in which detected blocks
appear as labeled groups, and a hierarchical view that draws one symbol per
block. The functional organization that published netlists leave implicit
(\S\ref{sec:intro}) thus becomes part of the released record, and the
generated views keep the entry legible to human designers.

The matcher draws on a library of hand-authored functional templates, one
small netlist per structural variant, grouped by functional family in
Table~\ref{tab:templates}. Current-mirror variants make up more than half the
library because a mirror's function is set almost entirely by its wiring
topology~\cite{analogdb_template_massierSizingRulesMethod2008,analogdb_template_aggarwalComparativeStudyVarious2016}.
The library is device-level at the archived revision. Common-mode
feedback, which every fully differential design requires,
enters the corpus by the representation's other route: as released
entries rather than as templates. The corpus carries a continuous-time
5T sense amplifier in each input polarity, a behavioral sense servo,
and a discrete-time switched-capacitor detector
(Table~\ref{tab:coverage}), and each of the five composites of
\S\ref{sec:composition} closes its common-mode loop by instantiating
one of them at a pinned design point. What no template yet recognizes
is that same structure sitting inline in a flat netlist, so the
coverage statistics of \S\ref{sec:coverage} attribute no role to it
(\S\ref{sec:limitations}).
Matching semantics bound what detection can claim, so this section
states them exactly. The map $\varphi$ of
Algorithm~\ref{alg:block-detection} is an injective homomorphism that
must be induced on the template's internal nets and boundary-free at its
terminals: an internal net of $t$ must map to a private host net of
equal degree. A template match therefore fails if any internal net
carries even one extra legitimate load, a strictness that keeps every
emitted role exact but forfeits near-miss variants, and is one reason
coverage stops at 51\% (\S\ref{sec:coverage}). Matched instances may
overlap on devices, as when one diode device anchors two mirror
instances, but each covered device receives a single recorded structural
role. Subgraph isomorphism is NP-complete in general; the template
graphs are small and the kind, polarity, and pin-multiset labels prune
the search, so matching the full corpus completes as an ordinary step
of release generation.
This deterministic, release-time annotation complements the
netlist-annotation line of
\cite{EDA_annotation_gana_kunalGANAGraphConvolutional2020,EDA_annotation_gnn_kunalGNNBasedHierarchical2023,EDA_annotation_libfree_neunerLibraryFreeStructure2021},
which recognizes structure in user-supplied netlists at use time.

\begin{table*}[t]
  \caption{Sub-block template library at the archived revision, grouped by
  functional family. The library is device-level throughout; no
  common-mode-feedback family exists at this revision
  (\S\ref{sec:topology}).}\label{tab:templates}
  \centering
  \begin{tabular}{@{}l p{8.4cm} r@{}}
    \toprule
    Group & Variants & Templates \\
    \midrule
    Current mirrors & simple, cascode, Wilson, improved Wilson, high-swing
    cascode, improved high-swing cascode, low-voltage cascode, self-biased
    high-swing cascode, wide-swing~\cite{analogdb_template_massierSizingRulesMethod2008,analogdb_template_aggarwalComparativeStudyVarious2016}
    & 18 \\
    Differential pairs & simple, cascoded~\cite{analogdb_template_massierSizingRulesMethod2008} & 4 \\
    Cross-coupled pairs & simple, negative-resistance core~\cite{analogdb_template_massierSizingRulesMethod2008} & 2 \\
    Pseudo-resistors\(^{\dagger}\) & series, well and generator each
    independent or shared~\cite{analogdb_template_guglielmiHighValueTunablePseudoResistors2020}
    & 4 \\
    Switches & bulk-blind, rail-tied-bulk transmission gate (complementary);
    differential chopper modulator (\(4\times\) rail-tied-bulk pairs, same
    templates) & 2 \\
    Inverter / push-pull stage & complementary CMOS stack & 1 \\
    \midrule
    Total & & 31 \\
    \bottomrule
  \end{tabular}
  \vspace{2pt}

  {\footnotesize \(^{\dagger}\)Partial catalog: the 4 series-symmetric cells
  of Fig.~3 in Guglielmi et al.\ are drawn and registered; the remaining 12
  topologies of their \S4 (single-cell transdiodes, parallel-symmetric
  cells, linearity extensions, floating-voltage generators) are not yet
  templated.\par}
\end{table*}

\subsubsection{Device Parameterization and Matching Constraints}
\label{sec:params}
Reuse needs more than named device sizes: it also needs the matching
intent that a flat netlist discards. The parameter space of~\eqref{eq:entry}
therefore has three components in two layers, $\Theta = (\Sigma,
\sim, \mathcal{R})$. A generated
layer assigns one symbolic variable to every device geometry field, the
symbol set $\Sigma$: a complete, mechanical inventory of the design's
degrees of freedom. An authored layer then declares how those symbols
relate. Tie groups induce the equivalence $\sim$, whose classes share a
value: a differential pair matches in full geometry; a mirror rail shares
its length. Ratio declarations $\mathcal{R}$ freeze designed current
gains as relations rather than free values.

Individual parameters can also be frozen without being removed from the
record. The tie-group representatives left free after ratios and freezes
define the search space an optimizer may explore, and each per-kit
binding assigns them a default value and a search band in that
technology. Matching
constraints are therefore stored with the design as explicit, reviewable
statements (R3); Appendix~\ref{app:dsl} (Listings~\ref{lst:params}
and~\ref{lst:sizing}) shows both layers on a two-stage amplifier.

\subsubsection{Executable Datasheets}
\label{sec:datasheet}
Published performance tables seldom include the procedure that measured
them; the datasheet makes that procedure part of the design (R2). The
contract $C$ of~\eqref{eq:entry} assigns each reported metric $m$ a triple
$(a_m, x_m, S_m)$: the analysis $a_m$ that produces it, the extraction
$x_m$ that reads a number out of the resulting waveforms, and the
specification band $S_m$ that separates pass from fail,
\begin{equation}
  \operatorname{pass}(m, b, c) \iff
  x_m\bigl(\operatorname{sim}_c \mathcal{L}(T, a_m, b)\bigr) \in S_m ,
  \label{eq:contract}
\end{equation}
where $c$ ranges over the corner selections the binding $b$ declares and
$\operatorname{sim}_c$ evaluates the deck at that corner. A design point
records one verdict per declared corner, and every design point at the
archived revision evaluates the typical corner
(\S\ref{sec:limitations}).
Analyses themselves instantiate
class-owned testbench templates at stated operating conditions, and the class
registry also owns the canonical metric vocabulary, so a phase margin
measured on one amplifier means the same bench, extraction, and conventions
as a phase margin measured on any other. Because the verification harness
and the results scoreboard both judge against this one artifact, a reported
number cannot disagree with the procedure that produced it.
One caveat governs every measurement the contract produces: the release
simulates nominal schematics with perfectly matched devices, so all
reported rejection, offset, and residual-ripple figures are
\emph{systematic CMRR/PSRR (mismatch excluded)}, the term used for
them throughout. They are upper bounds on silicon, where mismatch,
which the current release does not model (\S\ref{sec:limitations}),
sets these quantities.
Listings~\ref{lst:analysis} and~\ref{lst:datasheet} give the concrete form.

\subsubsection{Hierarchical Composition}
\label{sec:composition}
\begin{figure*}[t]
  \centering
  \subfloat[CCIA system\label{fig:hierarchy-system}]{%
    \includegraphics[width=0.30\textwidth]{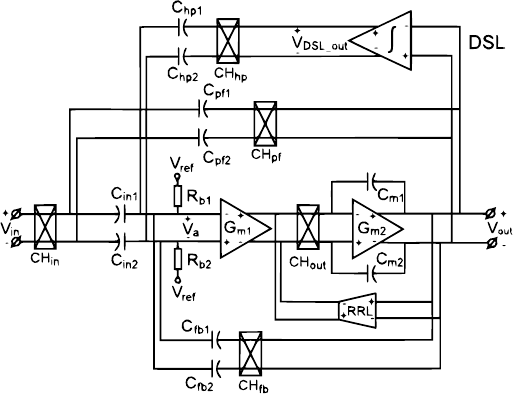}}
  \hfill
  \subfloat[Ripple-loop integrator OTA\label{fig:hierarchy-integrator}]{%
    \resizebox{0.64\textwidth}{!}{\includestandalone{media/F5b-annotated-integrator-switchcap-opamp}}}
  \par\vspace{4pt}
  \subfloat[Two-stage chopper core\label{fig:hierarchy-core}]{%
    \resizebox{0.825\textwidth}{!}{\includestandalone{media/F5c-annotated-two-stage-opamp-core}}}
  \caption{From blocks to systems: the chopper CCIA
  of~\cite{analogdb_design_chopper_ccia_fan18$mu$W602011} and its two
  op-amp cores, redrawn with functional groupings annotated. Groups
  matching a device-level template are matcher-detected: the
  differential pairs in (b) and (c) and the diode-referenced simple
  mirror in (b). The bias-source groups (current sources hung on the
  bias rails, with no diode reference in the drawn scope), the
  common-mode-feedback grouping, and the
  cascode-load grouping are designer annotations that the
  archived detection run does not emit (\S\ref{sec:limitations}). A device may sit in two
  groups (in (b), M15 is both the mirror reference and part of the
  CMFB branch) while carrying one recorded role
  (\S\ref{sec:topology}). (a)~The system with its capacitively coupled
  feedback loops; (b)~the ripple-reduction-loop integrator, a fully
  differential telescopic-cascode OTA with common-mode feedback;
  (c)~the two-stage chopper-stabilized core (folded-cascode $G_{m1}$,
  chopper, common-source $G_{m2}$).}\label{fig:hierarchy}
\end{figure*}
The step from blocks to systems is a composition manifest: an authored
declaration that builds a new circuit out of entries the database already
holds (R4). Each instance in the manifest names a block, pins the
exact revision of that block's topology by content hash, and wires its
ports; the child's released per-kit sizing supplies its geometry on
each kit the composite binds. The
composite may additionally expose an internal net of a block, remove a
block-local bias element it now supplies itself, or override a block
parameter; the instantiator owns the sizing of what it composes.

Generation then
flattens the manifest into an ordinary netlist and sizing binding, so
composition is closed over entries: a manifest $W$ over entries
$E_1, \dots, E_n$ yields
$\operatorname{compose}(E_1, \dots, E_n;\, W) = E'$, again of the
form~\eqref{eq:entry}. The composite binds exactly the kits every child binds: writing
$\mathrm{kits}(B)$ for the kits a binding set covers,
$\mathrm{kits}(B') = \bigcap_i \mathrm{kits}(B_i)$ (each child
instantiated at its pinned topology, with its released sizing on that
kit). A kit some child
does not bind is unavailable to the composite until that child gains
the binding. The composite therefore carries its own
datasheet, lowers to the same kits, and passes through the same verification
as any leaf circuit, and no downstream tool treats hierarchy as a special
case. Figure~\ref{fig:hierarchy} shows the pattern on the chopper
instrumentation amplifier
of~\cite{analogdb_design_chopper_ccia_fan18$mu$W602011}, whose system-level
entry composes the two op-amp cores shown. Among the corpus's five
composites, another
closes the common-mode loop of a fully differential two-stage amplifier
with a feedback servo drawn from a separate entry
(Listing~\ref{lst:composition}). A
sub-block published this way supports more than citation:
the verified block can be instantiated in a larger system at the design
point where it was verified.

\subsubsection{A Machine-Actionable Interface}
\label{sec:agent}
\begin{algorithm}[t]
  \caption{Agent loop over the file contract: discover, compare, reuse.}\label{alg:agent-loop}
  \begin{algorithmic}[1]
    \Require the released catalog; a target class, kit $k$, and constraints
    \State \emph{discover}: filter the catalog to entries of the class
      whose ports and kit bindings meet the constraints
    \State \emph{compare}: rank the candidates on their recorded design
      points on $k$: spec verdicts, power, area, headline metrics
    \State \emph{reuse}: select an entry $E = (T, \Theta, C, B)$ with
      binding $b_k \in B$ and fetch its released decks
      $\mathcal{L}(T, a, b_k)$
    \State simulate, or re-size the free symbols of $\Theta$ within the
      bounds of $b_k$ and simulate again
    \State judge every metric against the contract $C$; record the outcome
      as a new design point on the entry's scoreboard
  \end{algorithmic}
\end{algorithm}
An agent-facing corpus does not need a service; it needs a stable
interface. The interface to analog-db is a \emph{file contract}: the fixed
set of files each entry publishes, together with the schemas that govern
them, as distinct from the per-entry behavioral contract $C$
of~\eqref{eq:entry}. Every artifact
is governed by a versioned schema, and a generated catalog summarizes each
entry's class, ports, kit bindings, and recorded results in a single
queryable document, so any agent that can read files can use the database
without credentials, network access, or a running server (R5).
Algorithm~\ref{alg:agent-loop} states the loop this supports, and each pass
through it leaves a new
recorded design point behind for the next agent's comparison to draw on.
Append-only accession identifiers make every retrieved entry citable in
later work. The same contract also serves human readers, because the
generated views of \S\ref{sec:topology} keep every entry inspectable
without running any tooling. A protocol wrapper is deferred as future work
(\S\ref{sec:future}): a convenience layer over the same files rather
than new capability.

\subsection{Artifact Generation and Verification}
\label{sec:portability}
The representation of \S\ref{sec:representation} is only as credible as the
artifacts derived from it. This section states the guarantees the released
database makes: derived artifacts are reproduced mechanically from the
authored core, every entry carries a verification status with a defined
meaning, process neutrality is demonstrated by porting rather than
asserted, and what simulation measures is retained as recorded design
points rather than a single winner.

\label{sec:lowering}%
Lowering~\eqref{eq:lowering} resolves an entry's neutral topology through
one of its bindings into ready-to-run simulation decks, so a
reader reproduces a result by running a released deck rather than by
reassembling one. Retargeting synthesizes a new binding from an existing
one, extending the binding set $B$ of~\eqref{eq:entry} while $T$, $\Theta$,
and $C$ stay fixed: devices and corners are re-mapped onto the target kit
and sizings clamped to its geometry rules, after which the new binding
flows through the same lowering and verification path as any other.
Because a binding records only device and corner identifiers and
geometry rules, and never model files, the entry itself stays free of
process detail: what a new kit costs is a binding, not a rewrite.

\label{sec:verify}%
``Verified'' is a graded claim, and the database grades it explicitly.
Five tiers check progressively stronger properties. T0 checks schema
validity; T1 checks that every derived artifact regenerates byte for
byte; T2 assembles every circuit, analysis, kit, and corner combination
into a runnable deck; T3 simulates every released deck; and T4 checks
the measured metrics against the datasheet's specification bands, the
pass predicate of~\eqref{eq:contract}. The first three
tiers require no process kit and gate every change to the corpus. An
entry's published status is named for the highest tier it passes:
\emph{generated} (T1), \emph{simulated} (T3), \emph{validated} (T4);
\emph{validated} is earned only when every spec-bounded metric passes.
A baseline whose default sizing does not bias into a working circuit is
recorded as the starting point for a search rather than as
\emph{validated}, so \emph{validated} keeps its strict meaning. A separate structural check requires
every released schematic to be graph-isomorphic to the netlist re-derived
from it, so the generated views of \S\ref{sec:topology} can be trusted as
views of the same circuit.

\label{sec:scoreboard}%
Because an analog circuit meets a target application rather than a golden
specification (\S\ref{sec:intro}), the scoreboard names no scalar best.
Each simulation outcome can be recorded as a design point. For every
circuit and kit, the scoreboard marks the Pareto front over
power, area, and the headline metrics of the circuit's class, with one
named baseline per kit. Area is an acknowledged proxy: the summed gate
area of the sized MOS devices, with no spacing, routing, or well
overhead, and with capacitors and resistors excluded. The passive
exclusion matters: at typical
metal-insulator-metal densities of roughly 1.5--2\,fF/$\mu$m$^2$, a
single 10\,pF compensation capacitor occupies
5000--6700\,$\mu$m$^2$, comparable to or exceeding the recorded gate
area of many entries, so the proxy ranks device budgets rather than
die cost.
Because a new sizing extends the recorded tradeoff space instead of
overwriting a winner, the corpus can serve as an optimization
benchmark. Per-kit lookup tables for systematic
transconductance-efficiency ($g_m/I_D$)
sizing~\cite{analogdb_template_gmid_silveiraGmIDBasedMethodology1996,analogdb_template_gmid_jespersSystematicDesignAnalog2017} and
auto-generated optimizer configurations over the released
decks~\cite{nevergrad,bakshy2018ae} provide the means of producing new
design points, so an agent can size an entry against its own datasheet
without authoring any testbench
(exercised in \S\ref{sec:agent-eval} and Appendix~\ref{app:benchmarks}).
The tables follow the pre-computed lookup-table method of Jespers and
Murmann, whose published starter kits already sweep the same open
kits~\cite{general_Murmann_gmID_GitHub}, and are produced with an
open-source implementation of that flow~\cite{pygmid}; the archived
revision records the tables for the representative device pairs on
ihp-sg13g2 and sky130 and regenerates the remainder mechanically from
the released tools.

\label{sec:implementation}%
A single automation layer drives generation, verification, and
simulation. Every simulation in this release runs on an open-source
simulator (ngspice) against containerized open process kits, so the
whole flow is reproducible without a commercial license. Continuous
integration regenerates every derived artifact on each change and rejects
any drift, the mechanism behind the regeneration guarantee above.

\section{Experiments}
\label{sec:eval}
The evaluation asks five questions of the released artifact, each
answered by one subsection below: coverage and provenance (RQ1),
reproducibility (RQ2), portability across process kits (RQ3), whether
the agent-first interface supports design work rather than retrieval
alone (RQ4), and whether the recorded design points are dense enough to
expose performance tradeoffs and seed a sizing search (RQ5).

\subsection{Coverage and Provenance (RQ1)}
\label{sec:coverage}
\input{data/generated/coverage.tex}
\input{data/generated/class_status.tex}
RQ1 tests the claims behind the populated database and the sub-block
annotation scheme. At the archived revision the database holds 68
verifiable circuits across
sixteen classes: 33 OTAs/op-amps, 7 low-dropout regulators, 6
instrumentation amplifiers, and smaller classes that range from
common-mode feedback blocks to temperature sensors
(Table~\ref{tab:coverage}). Twenty-one of the 68
are fully differential, including four small blocks whose differential
outputs are declared in their netlists rather than in the catalog
summary. The count is conservative: it excludes three
incomplete entries and the 13 reference circuits retained for provenance
but never lowered. Table~\ref{tab:class-status} reports how far each
class has been carried through the verification tiers of
\S\ref{sec:verify}: 65 of the 68 entries hold at least one recorded
design point, 43 are validated on at least one kit, and 14 on all
three, so the corpus's verification depth is uneven across classes and
the table makes that unevenness inspectable rather than averaged away.

Provenance is part of each record. Every entry names the corpus or
collection it came from, and 48 of the 68 additionally cite the
published design they realize; the collection counts below sum to the
corpus, while the citation count is smaller because not every
collection entry realizes a single published design.
Twenty-two circuits are normalized from
the AnalogGym
corpus~\cite{analogdb_design_AnalogGym_liAnalogGymOpenPractical2024},
which contributes most of the three-stage OTAs along with regulators and
temperature sensors. Another 21 are hand-entered from published designs,
among them the chopper instrumentation-amplifier family of Fan et
al.~\cite{analogdb_design_chopper_ccia_fan18$mu$W602011}, the Hsu
biomedical front
end~\cite{analogdb_design_SRMC_iahsu68DBTHD2020,analogdb_design_classAB_output_IA_hsu18$mu$652018},
and the internal amplifiers of a commercial buffered-reference
regulator~\cite{analogdb_design_ldo_ti_texasinstrumentsTPS7A212022}.
Five circuits come from Pretl's teaching
collection~\cite{general_Pretl_AICD_GitHub}, one from
CORA-OpAmp~\cite{example_cora_opamp}, and five are composites of
\S\ref{sec:composition}, authored in this work. The remaining 14 are
drawn from other upstream sources or authored directly, with per-circuit
provenance and licensing recorded in the release's \texttt{NOTICE} file
(Appendix~\ref{sec:availability}).

How much of each entry the representation explains is itself measured
and released. Parameterization is complete: the
generated layer names every geometry field of the corpus's 1574 devices
as a symbolic variable (3819 symbols on the representative kit,
ihp-sg13g2), and the authored layer constrains that inventory: 1866
symbols are tied to a group representative through 444 tie groups, 255
are ratio-bound, and 365 are frozen,
leaving 1333 free symbols to search. Twenty-two circuits also record structural symmetries
left untied as reviewable warnings.
Structural detection, by contrast, is partial. All 68 circuits
carry generated structural views, spanning 1574 devices; the template
matcher of \S\ref{sec:topology} assigns a functional role to 807 of
them (51\%), organized into 321 detected block instances.
\S\ref{sec:limitations} records the uncovered remainder as a limit of
the current template library.

\subsection{Reproducibility (RQ2)}
\label{sec:repro}
RQ2 tests whether a reader can reproduce the released results from the
released artifacts. Reproducibility
rests on regeneration: simulation decks are derived mechanically from
the
authored core, and tier T1 of \S\ref{sec:verify} checks that they
regenerate byte for byte, so the deck a
reader runs is the one that produced the recorded design point. Two checks exercise this at
the archived revision. First, every value in the benchmark tables of
Appendix~\ref{app:benchmarks} is re-simulated from the archived
database, except where a table footnote records otherwise; none is
transcribed from a source publication. Second, a separate verifier audited the regulator study of
\S\ref{sec:agent-eval}, re-running sixteen recorded circuit-kit
baselines from the released sizing defaults rather than from the
sizing sessions' state, and reproduced every metric recorded for those
sixteen bit for bit.
Both checks demonstrate reproduction at the archived revision and
the typical corner.

\subsection{Cross-PDK Portability (RQ3)}
\label{sec:portability-eval}
RQ3 tests whether process neutrality (R1) holds in practice. The
low-dropout corpus is the systematic case: the released regulators
carry twenty-three circuit-kit bindings across the three open kits
(\texttt{ldo\_005} carries no sky130 binding, while \texttt{ldo\_007}
contributes three bindings without being counted among the seven
regulators of Table~\ref{tab:coverage}), and all twenty-three
pass their complete datasheets (Appendix~\ref{app:ldo-b}). That claim
is per-entry, judged against the per-entry specification bands of
Table~\ref{tab:ldo-specs}, which vary deliberately across entries: the
capacitor-less \texttt{ldo\_001} answers to looser
regulation bands than \texttt{ldo\_005}. Judged instead against the
single class-level reference band defined in
Table~\ref{tab:ldo-ppa}, 10 of the 23 bindings meet it. The two
verdicts answer different questions: conformance to each entry's
own contract versus performance against a band common to the class. The
benchmark
reports both rather than letting the first stand in for the second. No
port was mechanical, because an imported geometry encodes one
technology's assumptions. On gf180mcu the reference of
\texttt{ldo\_007} sat below a gate-source drop of the bound device
family. On sky130 a high-threshold device family drove a mirror diode
of \texttt{ldo\_004} to the rail at the released sizing's current density. The synthesized
\texttt{ldo\_005} binding first resolved to the kit's low-voltage
devices and had to be re-mapped to the high-voltage family that
matches its 3.3\,V rail.
What survives a port is the method. The detected sub-block roles of
\S\ref{sec:topology} state which devices each rule may touch, and
per-role \(g_m/I_D\) targets re-derive the geometry on each node
through that node's lookup tables. The tables also predict how close a port runs to its
limits: \texttt{ldo\_009}, which met a headroom bound on two
kits, showed a predicted onset of regulation within 30\,mV of the
simulated result. Each
port also adds to the entry's record: on \texttt{ldo\_005} the second
kit's leakage-free models exposed a latent defect that the first
kit's models had masked (Appendix~\ref{app:ldo-b}), and the repaired
netlist records the mechanism and the rejected repair as design notes
for the next retarget.

\subsection{Agent Case Studies (RQ4)}
\label{sec:agent-eval}
\begin{figure*}[t]
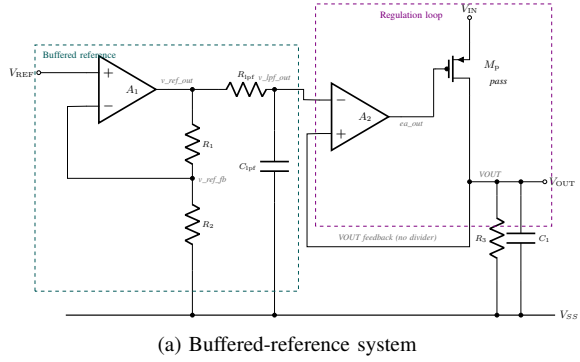
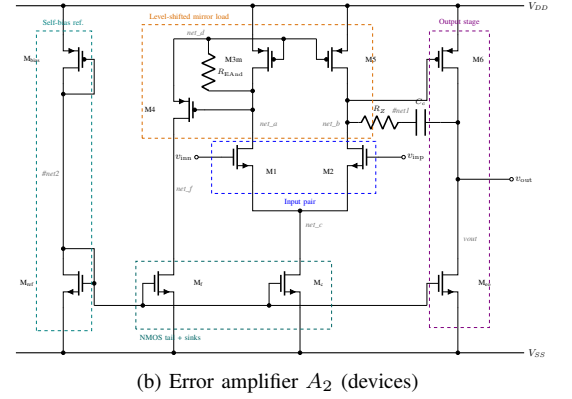

  \centering
  \subfloat[Buffered-reference system\label{fig:ldo005-system}]{%
    \resizebox{0.43\textwidth}{!}{\includestandalone{media/F5d-ldo-005-system}}}
  \hfill
  \subfloat[Error amplifier $A_2$ (devices)\label{fig:ldo005-erroramp}]{%
    \resizebox{0.40\textwidth}{!}{\includestandalone{media/F5e-annotated-ldo-005-error-amp}}}
  \caption{The verified \texttt{ldo\_005\_buffered\_ref}
  (Table~\ref{tab:ldo-ppa}), adapted from the TI
  TPS7A21~\cite{analogdb_design_ldo_ti_texasinstrumentsTPS7A212022}.
  (a)~The reference buffer $A_1$ and an RC low-pass filter feed the
  error amplifier $A_2$, which drives the PMOS pass device; the output
  is compared directly against the filtered reference, with no output
  divider. (b)~$A_2$, a self-biased two-stage Miller OTA with a
  level-shifted PMOS mirror load; the high-value tie $R_\mathrm{EAnd}$
  pins the otherwise floating mirror gate rail, the one-element repair
  of Appendix~\ref{app:ldo-b}. The commercial datasheet publishes only
  the functional block diagram; the transistor-level realizations of
  $A_1$ and $A_2$ are authored in this work, so the entry records an
  adapted re-realization, not the commercial part. In silicon the rail
  would be pinned by a diode-connected device or a dedicated bias
  branch; the resistive tie is the simulation-level equivalent, and
  the netlist notes record it as such.}\label{fig:ldo005}
\end{figure*}

RQ4 tests the agent-first interface of \S\ref{sec:agent} and whether
composition (R4) holds in use. We report two studies run as
author-supervised
sessions of a general-purpose large-language-model coding agent
(Claude Code, running the Claude Opus~4.8 model). The session records
identify the model only as Claude Opus~4.8 with the 1M-token context
window; we did not capture the exact dated model version
at the time, so the studies are reproducible in method but not
bit-for-bit in agent behavior.
In each study the
agent sized database entries to their datasheets using the released
artifacts and live
simulation on each bound kit. The first subject is a
buffered-reference low-dropout regulator bound to two process kits;
the second is the chopper instrumentation amplifier of
Fig.~\ref{fig:hierarchy}, carried from its op-amp cores to the closed
system. Success is defined by the database's own contract: a study is
complete when every metric of the entry's machine-readable datasheet
passes under the class benches. Each accepted result re-enters the
corpus as a recorded design point.

The supervision protocol was interactive rather than autonomous: the
author set each study's objective, reviewed the agent's proposed
diagnoses and edits, and accepted or rejected results against the
datasheet verdicts. Sizing values and the diagnostic probes that
located each failure mechanism originated with the agent. Each study
was recorded in a per-session engineering report carrying every
accepted number, root cause, and rejected repair. Neither those
reports nor the full conversational transcripts are part of the
release, and \S\ref{sec:limitations} records that gap. No quantified
per-intervention log was kept, so
the studies claim feasibility under supervision, not autonomy.

Each session ran the same loop over the released artifacts:
re-simulate the imported sizing, so that every later claim is a
delta against a reproducible baseline; localize the failure mechanism
through the detected sub-block roles, the per-kit \(g_m/I_D\) tables,
and targeted single-bench probes; re-size by converting per-role
\(g_m/I_D\) targets to geometry through the same tables; and re-judge
the full datasheet, accepting a trade only when the degraded metric
still meets its specification.

The regulator study applied this loop across the low-dropout corpus of
Appendix~\ref{app:ldo-b}: seventeen of the twenty-three imported sizings
failed their benches, and the loop closed each, typically within one to
three sizing iterations once a mechanism was identified. The
buffered-reference regulator
\texttt{ldo\_005}~\cite{analogdb_design_ldo_ti_texasinstrumentsTPS7A212022}
(Fig.~\ref{fig:ldo005}) concentrated the hardest failures: on
gf180mcu the best sizing that satisfied every other specification
still left a 519.5\,mV dropout against the 400\,mV bound, and the
port to ihp-sg13g2 failed outright. Probing below the sizing layer
exposed the shared cause, a netlist defect examined in
Appendix~\ref{app:ldo-b}: a mirror gate rail with no DC definition of
its own. A one-element repair that gives the rail a DC
definition (the tie of Fig.~\ref{fig:ldo005}), followed by a
compensation retune, superseded both verdicts: the regulator passes
all fourteen datasheet metrics on both kits (Table~\ref{tab:cs-ldo}),
with dropout at 340\,mV on gf180mcu and 170\,mV on ihp-sg13g2.

\begin{table}[t]
  \caption{Agent case study: the \texttt{ldo\_005\_buffered\_ref}
  datasheet specification against the released sizing on each bound kit
  (tt corner; 3.3\,V input, 1\,$\mu$F output capacitor, 1\,mA nominal
  load, per Table~\ref{tab:ldo-specs}). Every metric passes; the
  ihp-sg13g2 phase margin clears its bound by
  1.4$^\circ$.}\label{tab:cs-ldo}
  \centering
  \small
  \setlength{\tabcolsep}{4pt}
  \begin{tabular}{l l r r}
    \toprule
    Metric & Spec. & gf180mcu & ihp-sg13g2 \\
    \midrule
    Output voltage (V)               & 1.50--1.70 & 1.64  & 1.61 \\
    Quiescent current ($\mu$A)       & $\leq 800$ & 472.4 & 758.7 \\
    Load regulation (mV)             & $\leq 60$  & 6.0   & 1.3 \\
    Line regulation (mV)             & $\leq 300$ & 17.2  & 4.3 \\
    Dropout (mV)                     & $\leq 400$ & 340.4 & 169.7 \\
    PSRR at 1\,kHz (dB)              & $\geq 25$  & 35.6  & 44.5 \\
    $Z_{\mathrm{out}}$ peaking (dB)  & $\leq 10$  & 3.7   & 5.9 \\
    Load-step undershoot (mV)        & $\leq 150$ & 5.2   & 2.0 \\
    Load-step settling ($\mu$s)      & $\leq 30$  & 21.2  & 0.9 \\
    Output noise (mV$_{\mathrm{rms}}$) & $\leq 20$ & 7.5  & 5.0 \\
    Line-step ripple (mV$_{\mathrm{pp}}$) & $\leq 30$ & 9.2 & 3.4 \\
    Loop phase margin ($^\circ$)     & $\geq 45$  & 52.6  & 46.4 \\
    Loop gain (dB)                   & $\geq 50$  & 83.3  & 91.9 \\
    Loop unity-gain freq.\ (kHz)     & $\geq 100$ & 294.8 & 569.1 \\
    \bottomrule
  \end{tabular}
  \vspace{2pt}

  {\footnotesize The bands target functional closure of the adapted
  topology on each kit, not competitive power: the $\leq 800$\,$\mu$A
  quiescent-current band sits two orders of magnitude above the
  6.5\,$\mu$A typical of the commercial part the topology is adapted
  from~\cite{analogdb_design_ldo_ti_texasinstrumentsTPS7A212022}.\par}
\end{table}

The instrumentation-amplifier study began one stage earlier, at
bring-up: the system entered the database from hand-drawn schematics,
and characterization exposed defects that had survived drawing and
inspection. An input modulator gated the signal instead of chopping
it (roughly 1.6\,V/V with the clocks running against 19.3
with them static), and four device source terminals sat on floating nets, a defect
invisible on the printed schematic. Debugging then hinged on locating
mechanisms rather than searching over sizes: a ripple-sensing
integrator ran at the chop frequency rather than twice it, a
consolidation of clock phases silently inverted an impedance-boost
path into positive feedback, and beneath both lay a core that is
bistable without common-mode feedback. The agent closed the
common-mode loop with a servo whose injection point and drive limits
it chose by measurement.

The study ended with composition
(R4): on the open 130\,nm kit, composing the sized blocks into the
closed system of \S\ref{sec:composition} nulled the zero-input output
ripple from approximately 0.40\,V$_{\mathrm{pp}}$ to below
0.1\,$\mu$V$_{\mathrm{pp}}$ (Table~\ref{tab:ia-ihp}), qualitatively
consistent with the ripple reduction reported for the source
design~\cite{analogdb_design_chopper_ccia_fan18$mu$W602011}. That
residual sits at the numerical floor of a noiseless
transient rather than at a measurable output level, and the
composition closes the ripple loop and the common-mode loop at once,
so the nulling cannot be attributed to either loop alone
(Appendix~\ref{app:ia-b}). All
reported values are single-corner measurements; the composite's
common-mode loop is closed by an idealized servo; and on the open
kits the clocked entries verify through windowed transient extraction
(\S\ref{sec:limitations}).

A black-box baseline shows what one sizing-only search reached on the
same artifact. Handed the same closed system as a flat parameter space, a
Bayesian optimizer~\cite{bakshy2018ae} produced one feasible trial in
its first nineteen, reached 16 of 30 only after its bounds were
narrowed to the neighborhood of a hand-seeded design point
(Table~\ref{tab:cs-bo}), and had still not
beaten that seed on any objective when its trial budget ran out. The
binding constraint was the unregulated common mode, which no move
inside the searched sizing space repaired: the optimizer could only
spend its budget against it, whereas the agent diagnosed it and
changed the topology. The comparison is therefore asymmetric
(an autonomous sizing-only search against author-supervised sizing plus
topology edits),
and it locates where the failure lay without constituting
a controlled comparison of two search methods on one move set.

No ablation isolates which part of the representation produced that
difference; what the session records show is how each part was used.
The detected roles scoped each sizing rule to the devices it was
meant for, and the block hierarchy let a two-loop regulator and a
clocked system be debugged loop by loop.
The agent identified most root causes from device operating points in the
\(g_m/I_D\) tables before any re-size. The
recorded baselines made each reported improvement a delta against a
re-simulable released sizing, and the design notes on each entry record
those conclusions for later sessions. Nor are the per-role sizing
rules specific to the two systems reported here: the
re-sized entries marked throughout the benchmark tables of
Appendix~\ref{app:benchmarks} were closed by constrained
re-optimization combined with those same rules.

No like-for-like simulator-cost comparison is reported. Black-box
re-sizing consumed hundreds of recorded evaluations per circuit and
kit, while the agent loop closed each regulator binding within four
recorded full-suite runs; but the agent's single-bench diagnostic
probes were not logged and the supervision itself carries no cost
record, so the two recorded costs count different things, and a
per-evaluation table would understate the agent side.

\begin{table}[t]
  \caption{Black-box baseline on the closed instrumentation-amplifier
  system: Bayesian optimization over the flat sizing space, 49 trials
  across three narrowed searches, judged on the optimizer's own bench
  (closed-loop small-signal gain with the chopper clocks held
  transparent, 1.2\,V supply; gate area summed over the searched core
  devices only).}\label{tab:cs-bo}
  \centering
  \small
  \setlength{\tabcolsep}{4pt}
  \begin{tabular}{@{}l c c@{}}
    \toprule
    Search space & Dim. & Feasible trials \\
    \midrule
    Devices + bias inputs, bounds $0.3\times$--$3\times$ & 18 & 0 of 4 \\
    Devices only, bounds $0.3\times$--$3\times$ & 14 & 1 of 15 \\
    Devices only, bounds $0.6\times$--$1.7\times$ & 14 & 16 of 30 \\
    \bottomrule
  \end{tabular}
  \vspace{2pt}

  {\footnotesize A trial is feasible when it satisfies every constraint
  of the optimizer's problem; the best-scoring trial need not be
  feasible. The hand seed (the released sizing defaults, never
  evaluated as an optimizer trial) dominates the best trial on all three
  objectives: gain 25.713 vs.\ 25.611\,dB, power 263.4
  vs.\ 283.9\,$\mu$W, searched-device gate area 96.8
  vs.\ 127.2\,$\mu$m$^2$. The system row of Table~\ref{tab:ia-ihp} is a
  different measurement of a later sizing: clocked windowed-transient
  gain at the 5\,kHz chop, gate area over all 75 devices.\par}
\end{table}

\subsection{Scoreboard Coverage and Optimization Support (RQ5)}
\label{sec:pareto-eval}
\begin{figure*}[t]
  \centering
  \includegraphics[width=\linewidth]{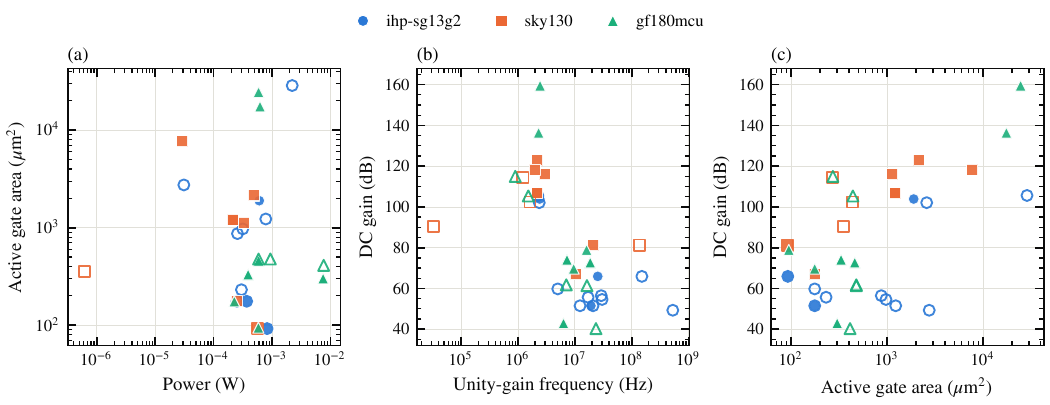}
  \caption{Validated amplifier design points, colored and shaped by
  process kit: (a) power against active gate area, (b) DC gain against
  unity-gain frequency, (c) DC gain against active gate area. Every
  plotted point passes its evaluated datasheet at the tt corner and
  each kit's default supply, under the bench conditions its record
  declares. Hollow markers are the 19 of 39 points recorded before the
  benches standardized on a uniform 10\,pF load, judged under the
  lighter native-load benches then in force
  (\S\ref{sec:pareto-eval}); filled markers are judged at the uniform
  load. Area is the gate-area proxy of
  \S\ref{sec:scoreboard}.}\label{fig:ppa-landscape}
\end{figure*}
\begin{figure*}[t]
  \centering
  \includegraphics[width=\linewidth]{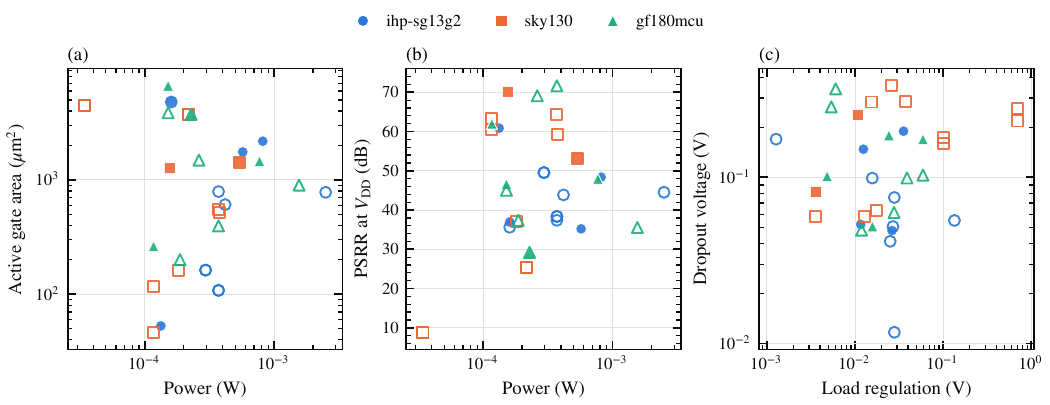}
  \caption{Validated regulator design points: (a) power against active
  gate area, (b) power-supply rejection at $V_{\mathrm{DD}}$ against
  power, (c) dropout voltage against load regulation. Every plotted
  point passes its evaluated datasheet at the tt corner, under the LDO
  class benches at each entry's datasheet operating conditions
  (Appendix~\ref{app:ldo-b}). Filled markers are the 10 of 34 points
  that also sit inside the class-level reference band of
  Table~\ref{tab:ldo-ppa}; hollow markers pass their own datasheet but
  fall outside it, or record too few band metrics to judge: all
  twenty-three baselines are judged, while the eleven unjudged points
  are alternative sizings predating the loop-break
  marker (Appendix~\ref{app:ldo-b}).}\label{fig:ppa-landscape-ldo}
\end{figure*}
RQ5 tests the optimization support the scoreboard of
\S\ref{sec:scoreboard} provides. At the archived
revision the record holds 457 design points, and 127 of the 159 populated
circuit-kit cells (one scoreboard cell per circuit and kit) carry two or
more; the per-cell Pareto marking
therefore discriminates among real alternatives. Sizings that miss
their datasheet stay in the record as starting points for a search.

Figure~\ref{fig:ppa-landscape} plots every validated design point
(\S\ref{sec:overview}) on the amplifier scoreboard, whereas the tables of
Appendix~\ref{app:amp-b} report only the single released sizing per
entry. Of the 327 recorded amplifier design
points at the archived revision, 39 are validated (the behavioral
\texttt{amp\_028} is excluded). The validated points span more than four decades of power
(0.6\,$\mu$W to 7.7\,mW), two and a half decades of active gate area,
and DC gains of 40 to 160\,dB at unity-gain frequencies from 33\,kHz to
528\,MHz, and the three open kits interleave across all three panels
rather than stratifying into per-kit bands.
Two caveats apply. First, each point is judged
under the bench conditions in force when it was recorded: the amplifier
benches are standardized to a uniform 10\,pF load at the archived
revision, but the scoreboard retains earlier points recorded against
lighter native-load benches, and the widest unity-gain frequencies,
including the 528\,MHz endpoint (a legacy-bench record of
\texttt{amp\_018}), are among those; the figure draws these 19
legacy-bench points hollow so the two populations stay
distinguishable. Second, the highest DC gains are
nominal schematic measurements of re-optimized gf180mcu sizings, and the
mechanism behind them is the re-optimizer driving channel lengths to the
geometry bound: gain per stage scales as
\((g_m/I_D)/\lambda\), and \(\lambda\) falls with \(L\), so on a 180\,nm
process a constrained search trades area for gain and phase margin. These
readings are physically reachable rather than erroneous, but they cost
area. Appendix~\ref{app:amp-b} marks the most extreme
of them against an absolute area threshold and names what that threshold
leaves unmarked, since the same mechanism operates below it.

Figure~\ref{fig:ppa-landscape-ldo} gives the regulator view, plotting
the 34 validated of the 71 recorded design points over power,
area, supply rejection, dropout, and load regulation. The panels
project onto regulation metrics because only one regulator's datasheet
declares a loop-gain bound at this revision, though the loop is now
measured on every binding. Both figures are generated
by script from the released catalog and scoreboard artifacts, so
any reader with the archived revision can regenerate them.

\section{Discussion}
\label{sec:discussion}
Taken together, the five research questions support one conclusion:
the obstacles to sharing analog designs are representational rather
than fundamental. Once a design is distributed as one
schema-governed object containing its topology, parameter space,
contract, and bindings, then reproducibility, portability, and machine
actionability are carried by the representation rather than by each
author's release practice.
The case studies indicate what changes as a consequence: design
knowledge that today survives only as a table in a paper becomes an
artifact a later designer or agent can re-run, re-size, and extend,
and each such use can leave a new recorded design point on the entry
it drew from. The limitations
below bound where this claim holds at the current release.

One methodological finding deserves prominence beyond its table
footnote: a loop phase margin certifies only the loop it measures.
On gf180mcu, \texttt{021\_yan\_az} sustains a 1.84\,V$_{\mathrm{pp}}$
self-starting oscillation with the input held at DC while its
outer-loop margin reads $+69^\circ$. The oscillator is internal to the
cell, where neither open-loop bench can see it. The corpus therefore treats a
zero-input transient as the deciding stability check, and this
case is why: a benchmark that certified stability by loop margin
alone would have published an oscillator as a passing entry.

\subsection{Limitations}
\label{sec:limitations}
Structural annotation is bounded by the template library at the
archived revision. The
matcher assigns a role to 51\% of the corpus's devices
(\S\ref{sec:coverage}) because structures outside that device-level
library contribute nothing: no common-mode-feedback template exists at
this revision (\S\ref{sec:topology}), and a common-mode loop composed
from a released entry is annotated only where its own devices match a
device-level template. On \texttt{amp\_029} and its CMFB-closed composite
\texttt{amp\_030} (\S\ref{sec:composition}) this leaves 38\% and 25\% of
devices with an assigned role, against 51\% across the corpus. Growing the
library and re-running detection is the first direction of
\S\ref{sec:future}.

The corpus is amplifier-weighted. Amplifiers and OTAs account for 33
of the 68 circuits, while the data-converter, sampler, and trim
classes hold one entry each (Table~\ref{tab:coverage}), so the
representation's generality beyond amplifier-like blocks is less
exercised than its depth on amplifiers.

The agent evidence is a pair of case studies, not a controlled
experiment. Both sessions were author-supervised runs of a single
coding agent and model, the supervision record is limited to
per-session engineering reports that are not themselves part of the
release, and the comparison arm that would isolate
the contribution of the structural artifacts---an agent denied those
artifacts---has not been run. The recorded simulator costs of the two
disciplines count different things (\S\ref{sec:agent-eval}), so no
cost comparison is tabulated. The optimizer
baseline is additionally asymmetric, sizing-only against sizing plus
topology edits (\S\ref{sec:agent-eval}). The studies
establish feasibility---that an agent working only from the released
artifacts can reach datasheet compliance---and attribute nothing
causally to any individual artifact.

Device mismatch is not modeled. Every reported figure comes from a
nominal schematic with perfectly matched devices, so the rejection,
offset, and residual-ripple values that mismatch sets in silicon are
systematic CMRR/PSRR (mismatch excluded) upper bounds
(\S\ref{sec:datasheet}), and readings at
the simulator's resolution floor are reported as bounds rather than
values (Appendix~\ref{app:amp-b}).

Recorded results are single-corner: every design point at
the archived revision was evaluated at the typical corner, and the
released $g_m/I_D$ tables are characterized there. Multi-corner PVT
evaluation and optimization remain future work.

Verification depth for clocked circuits is limited by the simulator.
Small-signal analysis is invalid for a chopper or switched-capacitor
circuit because it freezes the switches in one state, and the
open-source simulator (\S\ref{sec:implementation}) offers no periodic
steady-state alternative. On the open kits these entries fall back to
windowed transient analyses, leaving band-shape metrics as
specification targets rather than measured values
(Appendix~\ref{app:ia-b}), and no released entry yet carries a
periodic steady-state characterization. The clocked circuits are
therefore the least uniformly verified part of the corpus.

Finally, all verification is pre-layout. No entry carries a layout or
extracted parasitics, the recorded area is the acknowledged gate-area
proxy of \S\ref{sec:scoreboard}, and no released result has yet been
correlated against silicon.

\subsection{Future Work}
\label{sec:future}
Three directions follow. The first is detection coverage: adding a
common-mode-feedback family to the sub-block template library
(Table~\ref{tab:templates}) and growing the library further, starting
with the pending
pseudo-resistor variants, then re-running detection, would raise the
structural coverage reported in \S\ref{sec:coverage}. The second is hierarchy above the
sub-block level: an LLM annotation layer, cross-checked against the
deterministic matcher beneath it, could identify and verify
abstractions that subgraph matching cannot name, such as gain stages
and signal chains, and record them as queryable taxonomy metadata
alongside the structural roles. The third is breadth: new entries in
the thin classes, notably data converters, and advanced low-power OTA
topologies such as subthreshold and current-reuse designs, each
entering through the same schema and verification harness as the
existing corpus. Beyond these, the protocol wrapper of
\S\ref{sec:agent} remains future work: it would simplify
access over the same files rather than extend them.

\section{Conclusion}
\label{sec:conclusion}
This paper presented analog-db, a database built on a foundry-neutral
representation that captures each design as a topology, a set of
testbenches, and a machine-readable datasheet under one schema.
Entries annotate functional sub-blocks with parameters that carry their
matching constraints, and
verified blocks compose into larger systems. A file contract exposes
them to AI agents, while generated schematic views keep them readable
to designers. Derived decks regenerate byte-identically, every
entry carries a graded verification status, and the audited baselines
reproduce bit for bit at the archived revision.
Working only from the released artifacts, a supervised coding agent
carried an LDO
to full datasheet compliance on two kits. It assembled a published chopper
instrumentation amplifier from its op-amp cores into a closed-loop system,
with an idealized common-mode servo. In doing so it located four
defects introduced during hand entry, and a missing common-mode
feedback loop that the sizing-only baseline did not repair.
analog-db is released as a versioned, open corpus of 68 verifiable
circuits across sixteen classes. Every entry ships with its
testbenches, and 65 of the 68 carry recorded design points, so that
every reported result can be re-run and extended.

\section*{Acknowledgments}
This work was supported by the Natural Sciences and Engineering
Research Council (NSERC) of Canada through its Discovery Grant (DG)
Program under Grant \mbox{RGPIN-2024-06826}.

\bibliographystyle{IEEEtran}
\bibliography{references}

\appendices

\section{Benchmark Results}
\label{app:benchmarks}
\input{data/generated/ldo_spec_bands.tex}
This appendix reports the recorded design points behind the scoreboard
claims of \S\ref{sec:scoreboard}, one subsection per benchmarked class.
The landscape Tables~\ref{tab:amp-ppa}--\ref{tab:ia-ihp} are
gathered at the end of the paper. All
tabulated values are re-simulated from the released database at its
archived revision, except where a table footnote records otherwise;
none are transcribed from the source publications.

\subsection{Amplifier Benchmark}
\label{app:amp-b}

Table~\ref{tab:amp-ppa} reports the power, performance, and area of every
amplifier and OTA with an IHP 130\,nm (\texttt{ihp-sg13g2}) binding at its
released sizing. Tables~\ref{tab:amp-ppa-gf} and~\ref{tab:amp-ppa-sky} give
the same view for the GF180MCU and SKY130 bindings. Entries marked \(\S\)
do not retain their originally imported sizing. Stability verification
found the imported points electrically dead (no device biased into a
working operating region), unstable, or marginally stable, and they were
re-sized by constrained re-optimization combined with per-role
$g_m/I_D$ sizing. The table footnotes record the affected entries and the
failure each exhibited.
Throughout these tables, a circuit's per-kit rows are independently
sized realizations of one topology, not one design measured three
times, so power, gain, and area may differ widely across kits
(\texttt{025\_hsu\_classab\_ota} spans 13.1\,$\mu$W on sky130 to
1752\,$\mu$W on gf180mcu). Where a re-optimizer reached stability by
pushing geometry to its bounds, the row is marked as an optimizer
artifact rather than a defensible design (footnote of
Table~\ref{tab:amp-ppa-gf}).

That verification also uncovered a defect in the measurement itself
rather than in the circuits, one invisible to inspection of the recorded
values. Both open-loop benches referenced phase to the first swept
point, which is
correct only while that point lies below the dominant pole; at a
163.6\,dB plateau the dominant pole $f_p = \mathrm{UGF}/A_0$ sits in
the tens of millihertz, under the benches' 0.1\,Hz reference. Measured
on \texttt{006\_leung\_dfcfc1} on gf180mcu at the sizing then under
verification, the raw unwrapped phase at 0.1\,Hz sat $78.6^\circ$
below its 1\,$\mu$Hz value, and the self-reference subtracted that
accrued lag from every point including the crossover: the bench
reported $+72.7^\circ$ where the true margin was $-5.9^\circ$, and
read 150.0\,dB off the 163.6\,dB plateau. (The row
Table~\ref{tab:amp-ppa-gf} tabulates for this entry is the later
re-sized design point, at 1.32\,MHz.) The repair keeps the unwrapped
phase and its start-of-sweep reference and moves the sweep start below
the dominant pole ($10^{-4}$\,Hz on the affected high-gain cells),
where lowering it further does not change the reported margin. Six
entries were affected, all of them
high-gain multi-stage cells on gf180mcu, and the error is not a constant
offset: it tracks each amplifier's own $f_p$. Transient simulation
resolved the ambiguity: every entry the corrected reference drives
negative sustains a
limit cycle that self-starts with the input held at DC, while every entry it
leaves alone settles.
The corrected benches agree with the independent loop-gain bench on 13
of 14 comparable entries, where they had differed by up to
$89^\circ$. With the corrected reference in force, one of the 87
published amplifier circuit-kit baselines carrying a phase-margin cell
(\texttt{030\_miller\_cmfb\_composite} on ihp-sg13g2) sits below
the $45^\circ$ floor, at $33.3^\circ$. The floor itself is a
permissive bound (design practice and vendor datasheets more
commonly ask for $60^\circ$), so clearing it certifies stability of
the recorded point, not competitive phase margin. Two gf180mcu entries are withheld
from the tables entirely, and named
in their footnote, because no sizing was found for them that is both stable
and in specification: \texttt{011\_peng\_iac}, whose only stabilizing sizing
costs 37\,dB of PSRR, and \texttt{021\_yan\_az}, which self-oscillates at
1.84\,V$_{pp}$ held at DC and keeps oscillating with the outer loop
AC-opened, even as its outer-loop margin reads $+69^\circ$
(\S\ref{sec:discussion}). Both circuits remain in the database with their other kits and
their recorded measurements; what is withheld is the claim of a design
point, while the data remain.
The uniform 10\,pF load these tables share is a deliberate departure
from each source's own operating point, made for cross-cell
comparability: the AnalogGym shared amplifier testbench, for example,
loads all seventeen of its amplifier netlists with 500\,pF at a 1.8\,V
supply, so re-simulating those netlists at 10\,pF at each kit's own
supply moves multistage nested-Miller compensation designs $50\times$
from the load they were compensated for, and partly explains why many
imported sizings failed stability verification before re-sizing.
Table~\ref{tab:amp-native-loads} lists every cell evaluated $10\times$
or more from its as-published load; those rows are not comparable
to the source publication's numbers, and the native values are
retained so a reader can re-run any entry at its published operating
point.
\input{data/generated/amp_native_loads.tex}
The chopper rows record one architectural finding: the chopper OTA core
(\texttt{026}) carries no common-mode feedback and its systematic
CMRR/PSRR (mismatch excluded)
is correspondingly poor, while its CMFB-closed variant (\texttt{034})
recovers more than 40\,dB of both on ihp-sg13g2 for
less than 1\,dB of differential gain (footnote of
Table~\ref{tab:amp-ppa}).

\subsection{LDO Benchmark}
\label{app:ldo-b}

The top panel of Table~\ref{tab:ldo-ppa} reports the low-dropout
regulators with an ihp-sg13g2 binding, measured by the LDO class benches
against the per-entry specification bands and operating conditions of
Table~\ref{tab:ldo-specs}:
load and line regulation, dropout, power-supply rejection at 1\,kHz,
the loop phase margin (with the limits set out below), and a load-step
transient at each entry's own step. The class benches fix the operational
definitions: dropout is the smallest
$V_{\mathrm{in}}-V_{\mathrm{out}}$ at which the output is
simultaneously within 50\,mV of its target and decoupled from the
input ($|dV_{\mathrm{out}}/dV_{\mathrm{in}}| \leq 0.1$), so a pass
device that is off and merely follows the input cannot report a
near-zero dropout; load and line regulation are each the total output excursion
in volts (no slope normalization) across the full DC load and supply
sweeps;
the loop phase margin is taken by the \texttt{ac\_loopgain} bench under
Middlebrook voltage injection at a 0\,V break marker placed inside the
device under test, so it measures the regulation loop directly rather
than inferring it from the closed-loop output impedance;
load-step undershoot is measured against the post-step settled output
rather than the pre-step value; output noise is the output-referred rms
integrated over 10\,Hz--10\,MHz; and line-step ripple is the
peak-to-peak output deviation over a window spanning both supply edges,
so it includes the DC line-regulation shift between the two rails as
well as the transient excursions.

Through v1.1.1 this band judged stability on closed-loop
output-impedance peaking instead: the rise of $|Z_{\mathrm{out}}|$
above its 10\,Hz value over 10\,Hz--10\,MHz. That is a proxy rather
than a damping measurement, and it fails in two opposite directions. Referenced to its own
10\,Hz value, the reading tracks how far the regulation loop has already
rolled off by the frequency of the peak, which is a statement about loop
gain rather than about damping. And an entry whose output capacitor
holds $|Z_{\mathrm{out}}|$ under its low-frequency value across the
whole sweep reads 0\,dB whether its loop is well damped or barely
closed. The corpus splits on exactly that line: \texttt{001},
\texttt{002} and \texttt{003} are capacitor-less entries carrying
50\,pF, and all nine of their bindings fail the 10\,dB sub-band, which
is nine of the eleven failures; the remaining entries carry 1\,$\mu$F,
and seven of the twenty-three bindings sit on the 0.0\,dB floor
exactly.

Binding \texttt{ac\_loopgain} across the class was a queued database
update; it was applied in analog-db v1.1.2, so \texttt{pm\_loop\_deg} is now
recorded on all twenty-three baselines and the band of
Table~\ref{tab:ldo-ppa} is judged on it, against the 45$^\circ$ bound
\texttt{005} already carries in its own datasheet
(Table~\ref{tab:cs-ldo}). On the same pin the output-impedance proxy
puts 3 of 23 within the class band and the phase-margin band puts 10 of 23,
so the
shift is a change of measurement, not a change of corpus. The two disagree in the direction the construction
predicts. The nine capacitor-less bindings the proxy
penalized hardest measure 89.5$^\circ$ to 136.0$^\circ$ of phase
margin, among the best damped in the corpus, while the one binding that
fails the phase-margin sub-band, \texttt{008} on gf180mcu at
35.3$^\circ$, ranked only eleventh of twenty-three on the proxy at
12.5\,dB: flagged, but below all nine of them. The failures now
concentrate in $I_q$ rather than in a stability reading. The same sweep
yields the sensitivity-function peak
$\max |Z_{\mathrm{out,cl}}|/|Z_{\mathrm{out,ol}}|$, recorded as
\texttt{ms\_peak\_db} and reported but not judged, since no entry
carries an authored bound for it.

Eight entries appear against the seven counted in
Table~\ref{tab:coverage} because the benchmark includes \texttt{007},
which the coverage table excludes as incomplete. Seven of the eight
entries were re-sized (marked \(\P\)) after the imported sizings exhibited
under-damped or collapsing load-step responses
(\texttt{001}--\texttt{003}), a 559\,mV dropout (\texttt{007}), a dropout
bench invalidated by out-of-window regulation (\texttt{004}), and a
$-17$\,dB PSRR reading that traced to a failed DC solve rather than a real
supply path (\texttt{009}). Re-sizing followed the per-role
\(g_m/I_D\) loop of \S\ref{sec:agent-eval}, starting from each
circuit's detected sub-blocks (current mirrors, differential pairs, tail
sources).

The same procedure was then applied to the gf180mcu and sky130 panels of
Table~\ref{tab:ldo-ppa}, where the imported sizings failed in
technology-specific ways; the fixes from the first kit did not carry over
mechanically. The error amplifier of \texttt{007} latched off from
power-up on gf180mcu because its 0.6\,V reference sits below one
gate-source drop of that kit's 3.3\,V device family, so the earlier
104\,dB PSRR figure had measured supply coupling into a disconnected node.
\texttt{009} met the same headroom bound on both kits, with the predicted
onset of regulation matching the simulated result to within 30\,mV. The
buffered-reference regulator \texttt{005}~\cite{analogdb_design_ldo_ti_texasinstrumentsTPS7A212022} (Fig.~\ref{fig:ldo005}) exposed
a latent netlist defect: the gate rail of its level-shifted mirror load
was defined only by device leakage, which gf180mcu's BSIM4 models supply
and IHP's leakage-free PSP models do not, leaving a stable but incorrect
equilibrium that railed every DC sweep. A one-element repair---a resistive tie from the mirror drain to the gate
rail (50\,k$\Omega$ on gf180mcu, 500\,k$\Omega$ on ihp-sg13g2) that
carries no DC current by construction---together with a compensation
retune, made the circuit datasheet-compliant on both kits. The 340\,mV dropout on
gf180mcu falls inside the 400\,mV specification that sizing alone had
been unable to reach on the defective netlist. The root-cause record
for each repair is stored with the entry, as design notes in its
released netlist.

\subsection{Instrumentation-Amplifier / Bio-AFE Benchmark}
\label{app:ia-b}

The six instrumentation-amplifier entries derive from two published
designs: the chopper capacitively coupled instrumentation amplifier with
ripple-reduction loop of Fan et
al.~\cite{analogdb_design_chopper_ccia_fan18$mu$W602011} and the Hsu
biomedical analog front-end
chain~\cite{analogdb_design_SRMC_iahsu68DBTHD2020}. Table~\ref{tab:ia-ihp}
reports their figures and per-entry status at the released sizing on
\texttt{ihp-sg13g2}. Chopper entries are characterized with their clocks
running, through transient analyses at the 5\,kHz system chop
frequency, each measured over a window placed after that entry's own
settling transient, since the small-signal benches that freeze the
switches are invalid for the class (\S\ref{sec:limitations}). The input impedance is
derived from the average commutated input current, and the band-shape
metrics remain specification targets bound only where a
continuous-time bench is valid. The composite \texttt{ia\_004} exercises
the closure property of \S\ref{sec:composition} on a complete system,
composing the bare amplifier (\texttt{ia\_002}) with the
switched-capacitor ripple loop and an
ideal common-mode servo on the core's shared current-source gate. The
loop is clocked at twice the chop rate so its two-phase network
completes one sample-and-transfer cycle in each half-period of the
demodulated waveform; both half-periods contribute charge---the factor
of two in the loop's measured
$2(C_s/C_{\mathrm{int}})f_{\mathrm{chop}}$ transfer. Because the
release simulates matched devices, the up-modulated offset ripple this
loop removes in silicon is absent by construction here; the
$\sim$0.40\,V$_{\mathrm{pp}}$ zero-input ripple of the loop-less core
is recorded as chopping ripple, growing as $1/f_{\mathrm{chop}}$,
compounded by the unregulated output common mode, which is bistable
under the running clocks (a $\sim$0.7\,ms relaxation oscillation
through the capacitive network). Composing in the loop and the servo
together nulls it below 0.1\,$\mu$V$_{\mathrm{pp}}$, the numerical
floor of the noiseless transient; the composition closes both loops at
once, so the nulling is attributed to the pair, and the variant that
would separate them (\texttt{006}) binds only its DC operating point
at this revision. The supporting front-end blocks are held as
separate entries: the amplifier core OTAs appear in
Table~\ref{tab:amp-ppa} (\texttt{026}/\texttt{027}/\texttt{031}), and the
Hsu switched-capacitor common-mode feedback block reaches validated
status (\S\ref{sec:verify}) standalone at its 1\,kHz system clock.

\section{Domain-Specific Language}
\label{app:dsl}
This appendix presents the concrete syntax of the representation and the
design decisions behind its schemas, using one circuit from the corpus
as a running example: the two-stage Miller OTA \texttt{amp\_022\_fer\_two\_stage}, with
the composite \texttt{amp\_030\_miller\_cmfb\_composite} illustrating
hierarchy. Every artifact conforms to a versioned schema
(\texttt{spicexplorer/<kind>@1}) validated at Tier~0, and every excerpt
below is taken verbatim from the released database (elisions marked
\texttt{\ldots}). Throughout, the listings mark the authored/generated split
of \S\ref{sec:overview} in comment.

\subsection{Expressing Parametrization}
Parametrization splits the two layers of \S\ref{sec:params} across two
artifacts. The process-neutral \texttt{abstract/params.yaml} carries the
generated device-to-symbol map and the authored ties beneath it, where
\texttt{groups} declare structural matching and \texttt{ratios} freeze
multiplier relationships.

\begin{lstlisting}[language=yaml,caption={\texttt{amp\_022}:
\texttt{abstract/params.yaml}: generated device$\to$symbol map, authored tying.},
label={lst:params}]
schema: spicexplorer/params@1
devices:                  # GENERATED from abstract/topology.cgraph.json
  CC:  {Value: x_dut_cc_value}
  XM0: {l: x_dut_xm0_l, m: x_dut_xm0_m, w: x_dut_xm0_w}
  ...                     # one row per device (11 devices)
groups:                   # AUTHORED default tying, human-reviewed
- name: input_pair
  kind: matched_pair
  members: [XM0, XM1]
  tie: [w, l, m]
  description: PMOS differential input pair - full-geometry match
...
ratios:                   # frozen gains survive as ratios, not free knobs
- {param: m, ref: XM6, of: XM7, ratio: 17/3,
   description: Stage-2 current-source gain (legacy m=17 vs 3), frozen.}
\end{lstlisting}

The per-PDK \texttt{pdk/<pdk>/sizing.yaml} binding then supplies the
per-technology defaults and bands, and \texttt{freeze: true} is how a
parameter is withheld from the search without leaving the record.

\begin{lstlisting}[language=yaml,caption={\texttt{amp\_022}:
\texttt{pdk/ihp-sg13g2/sizing.yaml}: per-PDK defaults, bounds, freezes.},
label={lst:sizing}]
schema: spicexplorer/sizing@1
pdk: ihp-sg13g2
variables:
- {name: x_dut_xm0_w, description: "PMOS input pair width per finger
   (XM0; XM1 tied, m=5)", default: 8u, min: 0.18u, max: 50u, unit: m}
- {name: x_dut_cc_value, description: "Miller compensation cap (F)",
   default: 2e-12, min: 2e-13, max: 1e-11, unit: F}
- {name: x_dut_xm0_m, description: "Input pair fingers (frozen; XM1 tied)",
   default: 5, min: 1, max: 16, freeze: true, is_integer: true}
...
\end{lstlisting}

A simulated design point records the result: the scoreboard stores the same
symbols with concrete values, keyed by a \texttt{design\_id} that hashes
\{circuit, PDK, sizing\}.

\begin{lstlisting}[language=json,caption={A recorded design point
(\texttt{scoreboard/ihp-sg13g2/<design\_id>.json}, excerpt).},
label={lst:designpoint}]
{ "design_id": "9c0de4bdfb",
  "circuit": "amp_008_leung_nmcf", "pdk": "ihp-sg13g2",
  "parameters": { "sizing": {
      "CAPACITOR_0": "5p", "CURRENT_0_BIAS": 1.2393e-05,
      "x_dut_xm7_w": 5.8248e-07, ... } },
  "metrics": { "tt": { "pm_deg": {"value": 61.069, "spec": "pass"},
                       ... } } }
\end{lstlisting}

\subsection{Expressing Metrics and Testbenches}
The class registry (\texttt{\_shared/classes/amplifier/}) owns the canonical
metric vocabulary and the testbench \emph{templates}; a circuit's
\texttt{analyses/<id>.yaml} instantiates one template at
stated operating conditions. This keeps the corpus measurable with one vocabulary (the same
\texttt{pm\_deg} means the same bench everywhere) while each circuit states
only its numbers.

\begin{lstlisting}[language=yaml,caption={\texttt{amp\_022}:
\texttt{analyses/ac\_open\_loop.yaml}: an analysis instantiates a template.},
label={lst:analysis}]
schema: spicexplorer/analysis@1
id: ac_open_loop
template: ac_open_loop_biaswrap_ibias   # class-owned .spice template
description: "Bias-wrapped open-loop AC gain/phase."
params: {VDD: 1.5, VCM: 0.6, IBIAS: 20u, CL: 500f,
         FSTART: 1, FSTOP: 1G, PPD: 50}
flags: [bias_wrap]        # 1T-L / 1T-C wrap: DC-closed, AC-open loop
produces: [dc_gain_db, ugf_hz, pm_deg]
\end{lstlisting}

The \texttt{datasheet.yaml} declares, per canonical metric, where it comes from
(\texttt{analysis} + \texttt{extract}) and what passing means (\texttt{spec});
the verification harness and the scoreboard both judge against this one artifact.

\begin{lstlisting}[language=yaml,caption={\texttt{amp\_022}:
\texttt{datasheet.yaml} (excerpt): metric = analysis + extraction + spec.},
label={lst:datasheet}]
schema: spicexplorer/datasheet@1
default_conditions:
  supply: {unit: V, typical: 1.5}
  cload:  {unit: fF, typical: 500}
metrics:
  ...
  pm_deg:
    analysis: ac_open_loop
    extract: {meas: pm}
    spec: {min: 60.0, max: 90, unit: deg}
...
\end{lstlisting}

\subsection{Expressing PDKs and Simulators}
PDK knowledge lives in one registry per technology
(\texttt{\_shared/pdk/<pdk>.yaml}); corner \emph{libraries} stay out of the
repository and are referenced by file name and section, so the database clones
and verifies PDK-free through the kit-free tiers of \S\ref{sec:verify}, while
the simulation tiers gate on PDK presence. The
\texttt{sim\_engine} marker selects the simulator backend; every
released PDK is simulated with ngspice.

\begin{lstlisting}[language=yaml,caption={\texttt{\_shared/pdk/ihp-sg13g2.yaml}
(excerpt): engine marker, device classes, corners, geometry rules.},
label={lst:pdkreg}]
pdk: ihp-sg13g2
sim_engine: ngspice          # in-library router marker (open lane)
supply: {default: 1.5, unit: V}
devices:
  nmos: {lv: sg13_lv_nmos, hv: sg13_hv_nmos}
  pmos: {lv: sg13_lv_pmos, hv: sg13_hv_pmos}
corners:
  lib_file: cornerMOSlv.lib
  sections: [mos_tt, mos_ss, mos_ff, mos_sf, mos_fs]
geometry: {min_l: 0.13u, min_w: 0.18u, max_l: 10u, max_w: 10u}
...
\end{lstlisting}

Each circuit binding then maps its generic device kinds onto that registry and
names its corner selections. Retargeting a circuit to an additional process
kit requires
only these files plus \texttt{sizing.yaml}:

\begin{lstlisting}[language=yaml,caption={\texttt{amp\_022}: per-circuit PDK
binding: generic$\to$kit device map and corner selections.},label={lst:pdkbind}]
# pdk/ihp-sg13g2/devices.map.yaml
devices:
  nmos: {model: sg13_lv_nmos}
  pmos: {model: sg13_lv_pmos}
# pdk/ihp-sg13g2/corners.yaml
corners:
  tt: [{lib_file: cornerMOSlv.lib, section: mos_tt}]
  ...
default_corner: tt
\end{lstlisting}

\subsection{Expressing Hierarchy and Sub-blocks}
A composite circuit is an authored manifest
(\texttt{spicexplorer/composition@1}) over existing database entries: each
instance names a block \emph{and pins the exact revision of its
topology} (a content hash the compose validator checks), wires its ports,
and may \texttt{tap} internal nets, \texttt{omit} block-local bias elements, or
\texttt{override} a block's parameters. These are the composite-level adjustments of
\S\ref{sec:composition} in concrete syntax; the flattened netlist and sizing
binding the generation step emits are themselves generated, and checked by
T1 for byte-identical regeneration.

\begin{lstlisting}[language=yaml,caption={\texttt{amp\_030\_miller\_cmfb\_composite}:
\texttt{composition.yaml} (condensed): CMFB closure of \texttt{amp\_029} by a
behavioral servo block.},label={lst:composition}]
schema: spicexplorer/composition@1
of: amp_030_miller_cmfb_composite
instances:
- name: core
  block: amp_029_two_stage_miller_comp
  pin: 67c6e24e4137            # pinned topology revision (content hash)
  ports: {vinp: vinp, vinn: vinn, voutp: voutp, voutn: voutn,
          vdd: vdd, vss: vss}
  taps: {vbl: vbl_ctl}         # expose an internal bias net
  omit: [Vbl]                  # composite owns this bias now
- name: servo
  block: cmfb_001_ideal_rsense_servo
  pin: 9c8861186398
  ports: {vinp: voutp, vinn: voutn, vcmfb: vcmfb_raw,
          vref: vref_cm, vdd: vdd, vss: vss}
  override: {rout_val: 10k}    # gain-1 correction (range-referencing)
nets:                          # composite-level glue
- "Vrefcm vref_cm vss dc {vcm_ref}"
- "Vbl0 vbl_ctl vcmfb_raw dc {vbl0}"
sizing:                        # composite-owned knobs
  ...
\end{lstlisting}

\section{Artifact Availability}
\label{sec:availability}
analog-db is released publicly as the \texttt{analog-db/} component of the
\texttt{spicexplorer-release} repository,
\url{https://github.com/MacAnalog/spicexplorer-release}. Every result
reported here comes from release \texttt{v1.1.2} of that repository, the
archived revision referred to throughout.

\begin{landscape}
\input{data/generated/amp_ppa_ihp-sg13g2.tex}
\end{landscape}

\begin{landscape}
\input{data/generated/amp_ppa_gf180mcu.tex}
\end{landscape}

\begin{landscape}
\input{data/generated/amp_ppa_sky130.tex}
\end{landscape}

\begin{landscape}
\input{data/generated/ldo_ppa.tex}
\end{landscape}

\begin{landscape}
\begin{table}[p]
  \caption{Instrumentation-amplifier / bio-AFE benchmark in IHP
  130\,nm (\texttt{ihp-sg13g2}) at the released sizing of each entry:
  tt corner, 1.2\,V, 50\,fF load, 5\,kHz chop where
  clocked.}\label{tab:ia-ihp}
  \centering
  \footnotesize
  \setlength{\tabcolsep}{4pt}
  \resizebox{\linewidth}{!}{%
  \begin{tabular}{l r r r r r r l}
    \toprule
    Circuit & Gain (V/V) & $Z_{\mathrm{in}}$ (M$\Omega$) & Ripple (mV$_{\mathrm{pp}}$) & $V_{\mathrm{os}}$ res.\ ($\mu$V) & Power ($\mu$W) & Gate area ($\mu$m$^2$) & Notes \\
    \midrule
    \texttt{001\_hsu\_bandpass\_classab} & 19.7\(^{c}\) & 201.6\(^{c}\) & -- & 40.3\(^{c}\) & 10.1 & 78 & validated 8/8 \\
    \texttt{002\_fan\_chopper\_simple} & 7.0 & 11.8 & 402 & $<$0.1 & 175.7 & 103 & ripple target unmet (no ripple loop) \\
    \texttt{003\_fan\_chopper\_pf} & 6.9 & 23.8 & 220 & $<$0.1 & 175.7 & 105 & positive-feedback $Z_{\mathrm{in}}$ boost $\times$2.0 vs.\ \texttt{002} \\
    \texttt{004\_fan\_chopper\_rrl} & 19.1 & 6.4 & $<10^{-4}$\(^{r}\) & $<$0.1 & 217.9 & 191 & validated 5/5; RRL nulling + CM servo closed \\
    \texttt{005\_hsu\_pga\_ideal} & -- & -- & -- & -- & -- & -- & ideal-amp PGA, eight-code sweep recorded\(^{p}\) \\
    \texttt{006\_fan\_chopper\_cmfb} & -- & -- & -- & -- & 186.0 & 138 & CM-closed \texttt{002}; DC operating point only on this kit\(^{m}\) \\
    \bottomrule
  \end{tabular}}
  \vspace{2pt}

  {\footnotesize \(^{c}\)Continuous-time entry (no chopper): closed-loop
  AC gain, static differential $Z_{\mathrm{in}}$ at 100\,Hz, and static
  $V_{\mathrm{os}}$.
  \(^{r}\)Zero-input residual ripple with the ripple-reduction loop
  active: the deterministic ripple nulled to the numerical floor of a
  noiseless transient, against $\sim$0.40\,V$_{\mathrm{pp}}$ for
  \texttt{002}, which lacks the loop.
  \(^{p}\)Eight-code validation is recorded (gains $-0.08$ to
  $+17.98$\,dB, THD 0.0018\,\%), but those benches are not bound to the
  scoreboard.
  \(^{m}\)Binds only the DC operating point that proves the common-mode
  loop closed (520\,mV against the core's unregulated 610\,mV); the
  remaining benches stay on \texttt{002} until the variant is
  sized. The chopped gain of \texttt{002}/\texttt{003} is a windowed
  transient average.\par}

  {\footnotesize The rows are unoptimized reimplementations of the source
  topologies~\cite{analogdb_design_chopper_ccia_fan18$mu$W602011,
  analogdb_design_classAB_output_IA_hsu18$mu$652018,
  analogdb_design_SRMC_iahsu68DBTHD2020} on \texttt{ihp-sg13g2}, not
  measurements of the published silicon, which is why the tabulated
  powers sit well above the 1.8--6.3\,$\mu$W the sources
  report.\par}
\end{table}
\end{landscape}

\end{document}

%% file: media/F5b-annotated-integrator-switchcap-opamp.tex
\ctikzset{tripoles/mos style/arrows, transistors/scale=1}
\definecolor{netgray}{gray}{0.45}
\tikzset{
  netlbl/.style={text=netgray, font=\scriptsize\itshape, inner sep=1pt},
  devlbl/.style={font=\scriptsize, inner sep=1pt},
}
\begin{circuitikz}[font=\small, line width=0.7pt]

  \draw[line width=1pt] (0.4,11) -- (19.3,11) node[right]{$V_{DD}$};
  \draw[line width=1pt] (0.4,1.0) -- (19.3,1.0) node[right]{$V_{SS}$};

  \node[pmos, xscale=-1] (M1) at (5,10) {};
  \draw (M1.S) -- (5,11);
  \draw (M1.D) -- (5,9.2);
  \node[devlbl] at (4.4,10) {M1};

  \node[pmos]             (M2) at (3,8.4) {};
  \node[pmos, xscale=-1]  (M3) at (7,8.4) {};
  \draw (M2.S) -- (3,9.2) -- (7,9.2) -- (M3.S);      %
  \node[netlbl] at (5,9.45) {net1};
  \node[circ] at (5,9.2){};                          %
  \draw (M2.G) -- ++(-0.7,0) node[ocirc]{} node[left]{$v_{in+}$};
  \draw (M3.G) -- ++(0.7,0)  node[ocirc]{} node[right]{$v_{in-}$};
  \node[devlbl] at (3.6,8.4) {M2};
  \node[devlbl] at (6.4,8.4) {M3};

  \node[pmos, xscale=-1] (M9)  at (3,6.4) {};
  \node[pmos]            (M10) at (7,6.4) {};
  \draw (M2.D)  -- (M9.S)  node[netlbl,pos=0.5,left]{net6};
  \draw (M3.D)  -- (M10.S) node[netlbl,pos=0.5,right]{net5};
  \draw (M9.G) -- (M10.G);                      %
  \node[fill=white,inner sep=1.5pt] at (5,6.4) {$V_{b1}$};
  \node[devlbl] at (2.4,6.4) {M9};
  \node[devlbl] at (7.6,6.4) {M10};

  \node[nmos, xscale=-1](M11) at (3,4.4) {};
  \node[nmos]           (M12) at (7,4.4) {};
  \draw (M9.D)  -- (M11.D);          %
  \draw (M10.D) -- (M12.D);          %
  \draw (M11.G) -- (M12.G);                     %
  \node[fill=white,inner sep=1.5pt] at (5,4.4) {$V_{b2}$};
  \node[devlbl] at (2.4,4.4) {M11};
  \node[devlbl] at (7.6,4.4) {M12};
  \coordinate (voutn) at (3,5.4);
  \coordinate (voutp) at (7,5.4);
  \node[circ] at (voutn){};
  \node[circ] at (voutp){};
  \draw (voutn) -- (1.0,5.4) node[ocirc]{} node[left]{$v_{on}$};
  \draw (voutp) -- (8.6,5.4) node[ocirc]{} node[right]{$v_{op}$};

  \node[nmos, xscale=-1](M13) at (3,2.4) {};
  \node[nmos, xscale=-1](M14) at (7,2.4) {};
  \draw (M11.S) -- (M13.D) node[netlbl,pos=0.5,right]{net7};
  \draw (M12.S) -- (M14.D) node[netlbl,pos=0.5,left]{net8};
  \draw (M13.S) -- (3,1.0);
  \draw (M14.S) -- (7,1.0);
  \node[devlbl] at (2.4,2.4) {M13};
  \node[devlbl] at (6.3,2.4) {M14};
  \draw (M13.G) -- ++(0,-0.8) coordinate (g13);
  \draw (M14.G) -- ++(0,-0.8) coordinate (g14);
  \draw (g13) -- (6.8,1.6)   (7.2,1.6) -- (g14);   %
  \draw (M14.G) -- (11.7,2.4) -- (11.7,5.4) -- (13.0,5.4);  %
  \node[circ] at (M14.G){};
  \node[circ] at (13.0,5.4){};                             %

  \node[font=\small\itshape] at (5,0.3) {Main OTA (telescopic cascode)};

  \node[pmos] (M4) at (14.0,10) {};
  \draw (M4.S) -- (14.0,11);
  \node[devlbl] at (14.6,10) {M4};
  \draw (M1.G) -- (M4.G);                        %
  \node[fill=white,inner sep=1.5pt] at (9.5,10) {$V_{b3}$};

  \node[pmos]            (M5) at (11.0,7.8) {};  %
  \node[pmos, xscale=-1] (M6) at (13.0,7.8) {};  %
  \node[pmos]            (M8) at (15.0,7.8) {};  %
  \node[pmos, xscale=-1] (M7) at (17.0,7.8) {};  %
  \draw (11.0,8.8) -- (17.0,8.8);                 %
  \draw (M4.D) -- (14.0,8.8);
  \draw (M5.S) -- (11.0,8.8);
  \draw (M6.S) -- (13.0,8.8);
  \draw (M8.S) -- (15.0,8.8);
  \draw (M7.S) -- (17.0,8.8);
  \node[netlbl] at (12.0,9.05) {net2};
  \draw (M5.G) -- ++(-0.55,0) node[left]{$v_{on}$};    %
  \draw (M7.G) -- ++(0.55,0)  node[right]{$v_{op}$};   %
  \draw (M6.G) -- (M8.G);                              %
  \node[fill=white,inner sep=1.5pt] at (14.0,7.8) {$v_\mathrm{ref}$};
  \node[devlbl] at (11.6,7.8) {M5};
  \node[devlbl] at (12.4,7.8) {M6};
  \node[devlbl] at (15.6,7.8) {M8};
  \node[devlbl] at (16.4,7.8) {M7};

  \draw (M5.D) -- (11.0,6.1);
  \draw (M7.D) -- (17.0,6.1);
  \draw (11.0,6.1) -- (17.0,6.1);    %
  \node[nmos] (M16) at (17.0,3.4) {};
  \draw (17.0,6.1) -- (M16.D);
  \node[circ] at (17.0,6.1){};                %
  \draw (M16.G) |- (17.0,4.8);                %
  \node[circ] at (17.0,4.8){};                %
  \draw (M16.S) -- (17.0,1.0);
  \node[netlbl] at (16.0,5.85) {net3};
  \node[devlbl] at (17.6,3.4) {M16};

  \draw (M6.D) -- (13.0,6.7);
  \draw (M8.D) -- (15.0,6.7);
  \draw (13.0,6.7) -- (15.0,6.7);      %
  \node[nmos] (M15) at (13.0,3.4) {};
  \draw (13.0,6.7) -- (M15.D);         %
  \node[circ] at (13.0,6.7){};         %
  \draw (M15.G) |- (13.0,4.8);                %
  \node[circ] at (13.0,4.8){};                %
  \draw (M15.S) -- (13.0,1.0);
  \node[netlbl] at (15.5,6.7) {net4};
  \node[devlbl] at (13.6,3.4) {M15};

  \node[font=\small\itshape] at (14.0,0.3) {Common-mode feedback bias};

  \node[circ] at (5,11){};    \node[circ] at (14,11){};
  \node[circ] at (3,1){};     \node[circ] at (7,1){};
  \node[circ] at (13,1){};    \node[circ] at (17,1){};
  \node[circ] at (13,8.8){};  \node[circ] at (14,8.8){};  \node[circ] at (15,8.8){};

  \draw[dashed, blue] (2.0,5.5) rectangle (8.0,9.3);
  \node[font=\scriptsize, text=blue, anchor=south west] at (2.1,5.55) {Cascode diff pair (M2,M3,M9,M10)};

  \draw[dashed, orange!85!black] (2.0,1.5) rectangle (8.0,5.3);
  \node[font=\scriptsize, text=orange!85!black, anchor=north west] at (2.1,5.25) {Cascode load (M11,M12,M13,M14)};

  \draw[dashed, violet] (4.0,9.1) rectangle (15.0,10.95);
  \node[font=\scriptsize, text=violet, anchor=north west] at (4.1,10.9) {PMOS bias sources (M1,M4)};

  \draw[dashed, teal] (2.0,1.5) rectangle (8.0,3.3);
  \draw[dashed, teal] (12.0,2.5) rectangle (14.0,4.3);
  \node[font=\scriptsize, text=teal, anchor=west] at (14.15,4.6) {Simple mirror (M13,M14,M15)};

  \draw[dashed, magenta!70!black] (10.0,2.5) rectangle (18.0,8.7);
  \node[font=\scriptsize, text=magenta!70!black, anchor=north west] at (10.1,8.65) {CMFB circuit (M5,M6,M7,M8,M15,M16)};

\end{circuitikz}

%% file: media/F5c-annotated-two-stage-opamp-core.tex
\ctikzset{tripoles/mos style/arrows, transistors/scale=1}
\definecolor{netgray}{gray}{0.45}
\tikzset{
  netlbl/.style={text=netgray, font=\scriptsize\itshape, inner sep=1pt},
  devlbl/.style={font=\scriptsize, inner sep=1pt},
}
\begin{circuitikz}[font=\small, line width=0.7pt]

  \draw[line width=1pt] (0.4,12.0) -- (25.4,12.0) node[right]{$V_{DD}$};
  \draw[line width=1pt] (0.4,1.0)  -- (25.4,1.0)  node[right]{$V_{SS}$};

  \draw[dashed, gray!60] (0.9,0.6) rectangle (15.6,11.55);
  \node[font=\small\itshape] at (8.2,12.3) {First stage ($G_{m1}$) -- folded cascode};
  \draw[dashed, gray!60] (18.5,0.6) rectangle (25.9,11.55);
  \node[font=\small\itshape] at (21.85,12.3) {Second stage ($G_{m2}$) -- CS stage};

  \draw (2.2,11.0) -- (24.2,11.0);
  \draw (2.2,11.0) node[ocirc]{} node[left]{$V_{b4}$};

  \node[pmos] (M1) at (5.0,10.3) {};
  \draw (M1.S) -- (5.0,12.0);
  \draw (M1.G) |- (4.7,11.0);
  \node[devlbl] at (5.6,10.3) {M1};

  \coordinate (net1_junction) at (5.0,9.1);
  \node[pmos]            (M2) at (3.18,8.2) {};
  \node[pmos, xscale=-1] (M3) at (6.82,8.2) {};
  \draw (M2.S) -- (3.18,9.1) -- (6.82,9.1) -- (M3.S);
  \draw (M1.D) -- (5.0,9.1);
  \node[netlbl] at (5.75,8.85) {net1};
  \draw (M2.G) -- ++(-0.7,0) node[ocirc]{} node[left]{$v_{inp}$};
  \draw (M3.G) -- ++(0.7,0)  node[ocirc]{} node[right]{$v_{inn}$};
  \node[devlbl] at (3.58,7.55) {M2};
  \node[devlbl] at (6.37,7.55) {M3};
  \node[circ] at (net1_junction) {};

  \node[pmos]            (M7) at (11.6,10.3) {};
  \node[pmos, xscale=-1] (M6) at (14.4,10.3) {};
  \draw (M7.S) -- (11.6,12.0);
  \draw (M6.S) -- (14.4,12.0);
  \draw (M7.G) |- (11.3,11.0);
  \draw (M6.G) |- (14.7,11.0);
  \node[devlbl] at (12.2,10.3) {M7};
  \node[devlbl] at (13.8,10.3) {M6};

  \node[pmos, xscale=-1] (M11) at (11.6,8.2) {};
  \node[pmos]            (M10) at (14.4,8.2) {};
  \draw (M7.D) -- (M11.S) node[netlbl,pos=0.5,right]{net4};
  \draw (M6.D) -- (M10.S) node[netlbl,pos=0.5,left]{net5};
  \draw (M11.G) -- (M10.G);
  \draw (13.0,8.2) -- (13.0,7.75) node[ocirc]{} node[right]{$V_{b3}$};
  \node[devlbl] at (11.0,8.2) {M11};
  \node[devlbl] at (15.0,8.2) {M10};

  \node[nmos, xscale=-1] (M12) at (11.6,5.6) {};
  \node[nmos]            (M13) at (14.4,5.6) {};
  \draw (M11.D) -- (M12.D);
  \node[netlbl] at (12.5,6.85) {net6};
  \draw (M10.D) -- (M13.D);
  \node[netlbl] at (15.95,7.4) {net7};
  \draw (M12.G) -- (M13.G);
  \draw (13.0,5.6) -- (13.0,5.15) node[ocirc]{} node[below]{$V_{b2}$};
  \node[devlbl] at (11.0,5.6) {M12};
  \node[devlbl] at (15.0,5.6) {M13};

  \node[nmos, xscale=-1] (M4) at (3.18,2.8) {};
  \node[nmos]            (M5) at (6.82,2.8) {};
  \draw (M4.S) -- (3.18,1.0);
  \draw (M5.S) -- (6.82,1.0);
  \draw (M4.G) -- (M5.G);
  \draw (5.0,2.8) -- (5.0,2.35) node[ocirc]{} node[below]{$V_{b1}$};
  \node[devlbl] at (2.58,2.8) {M4};
  \node[devlbl] at (7.42,2.8) {M5};

  \coordinate (net2_junction) at (3.18,4.9);
  \draw (M2.D) -- (net2_junction) -| (M12.S);
  \draw (net2_junction) -- (M4.D);
  \node[netlbl, left=3pt] at (net2_junction) {net2};
  \node[circ] at (net2_junction) {};
  
  \coordinate (net3_junction) at (6.82,4.3);
  \draw (M3.D) -- (net3_junction) -| (M13.S);
  \draw (net3_junction) -- (M5.D);
  \node[circ] at (net3_junction) {};
  \node[netlbl, left=3pt] at (net3_junction) {net3};

  \draw[dashed, violet] (4.05,9.35) rectangle (25.15,11.4);
  \node[font=\scriptsize, text=violet, anchor=south west] at (4.15,9.4) {PMOS bias sources (M1,M6--M9)};

  \draw[dashed, orange!85!black] (10.6,7.15) rectangle (15.4,11.25);
  \node[font=\scriptsize, text=orange!85!black, anchor=south west] at (10.7,7.2) {PMOS cascode (M6,M7,M10,M11)};

  \draw[dashed, blue] (2.3,7.0) rectangle (7.7,9.2);
  \node[font=\scriptsize, text=blue, anchor=south west] at (2.4,7.05) {Diff pair (M2,M3)};

  \draw[dashed, teal] (2.3,1.85) rectangle (7.7,3.75);
  \node[font=\scriptsize, text=teal, anchor=south] at (5.0,3.82) {NMOS bias sources (M4,M5)};

  \draw (16.9,5.6) rectangle (18.1,7.6);
  \draw (16.9,5.6) -- (18.1,7.6);
  \draw (16.9,7.6) -- (18.1,5.6);

  \coordinate (net6_tap) at (11.6,6.6);
  \coordinate (net7_tap) at (14.4,7.2);
  \draw (net6_tap) -- (14.4,6.6) -- (16.9,6.6);
  \draw (net7_tap) -- (16.9,7.2);

  \node[pmos] (M8) at (21.2,10.3) {};
  \node[pmos] (M9) at (24.2,10.3) {};
  \draw (M8.S) -- (21.2,12.0);
  \draw (M9.S) -- (24.2,12.0);
  \draw (M8.G) |- (20.9,11.0);
  \draw (M9.G) |- (23.9,11.0);
  \node[devlbl] at (21.8,10.3) {M8};
  \node[devlbl] at (24.8,10.3) {M9};

  \node[nmos] (M14) at (21.2,6.0) {};
  \node[nmos] (M15) at (24.2,6.0) {};
  \draw (M8.D) -- (M14.D);
  \draw (M9.D) -- (M15.D);
  \draw (M14.S) -- (21.2,1.0);
  \draw (M15.S) -- (24.2,1.0);
  \node[devlbl] at (21.8,6.0) {M14};
  \node[devlbl] at (24.8,6.0) {M15};

  \coordinate (voutp) at (21.2,8.35);
  \coordinate (voutn) at (24.2,8.35);
  \node[circ] at (voutp) {};
  \node[circ] at (voutn) {};
  \draw (voutp) -- (21.2,8.75) -- (25.4,8.75) node[ocirc]{} node[right]{$v_\mathrm{outp}$};
  \draw (voutn) -- (24.2,7.95) -- (25.4,7.95) node[ocirc]{} node[right]{$v_\mathrm{outn}$};

  \draw (18.1,6.0) -- (M14.G);
  \node[netlbl] at (18.9,5.7) {net8};
  \draw (18.1,7.2) -- (19.0,7.2) -- (19.0,4.5) -- (23.2,4.5) -- (23.2,6.0) -- (M15.G);
  \node[netlbl] at (20.4,4.25) {net9};

  \draw (20.2,6.0) to[C, l=\scriptsize Cm1] (20.2,8.35);
  \draw (20.2,8.35) -- (voutp);

  \draw (23.2,6.0) to[C, l=\scriptsize Cm2] (23.2,8.35);
  \draw (23.2,8.35) -- (voutn);

  \node[circ] at (5,12){};     \node[circ] at (11.6,12){};  \node[circ] at (14.4,12){};
  \node[circ] at (21.2,12){};  \node[circ] at (24.2,12){};
  \node[circ] at (3.18,1){};   \node[circ] at (6.82,1){};
  \node[circ] at (21.2,1){};   \node[circ] at (24.2,1){};
  \node[circ] at (4.02,11){};  \node[circ] at (10.62,11){}; \node[circ] at (15.38,11){};
  \node[circ] at (20.22,11){}; \node[circ] at (23.22,11){};
  \node[circ] at (5,2.8){};    \node[circ] at (13,5.6){};   \node[circ] at (13,8.2){};
  \node[circ] at (net6_tap){}; \node[circ] at (net7_tap){};
  \node[circ] at (20.2,6){};   \node[circ] at (23.2,6){};
  \node[circ] at (21.2,8.75){};\node[circ] at (24.2,7.95){};

\end{circuitikz}

%% file: data/generated/coverage.tex
\begin{table}[t]
  \caption{Circuit coverage by class.}\label{tab:coverage}
  \centering
  \begin{tabular}{lr}
    \toprule
    Class & Circuits \\
    \midrule
    Amplifiers / OTAs             & 33 \\
    Low-dropout regulators (LDO)  & 7 \\
    Instrumentation amplifiers    & 6 \\
    Support blocks                & 5 \\
    Common-mode feedback (CMFB)   & 4 \\
    Switches                      & 3 \\
    Temperature sensors           & 3 \\
    Comparators                   & 2 \\
    ADC                           & 1 \\
    Line driver                   & 1 \\
    Sampler                       & 1 \\
    Trim                          & 1 \\
    Voltage reference             & 1 \\
    \midrule
    Total (verifiable)            & 68 \\
    \bottomrule
  \end{tabular}
  \vspace{2pt}

  {\footnotesize The Support blocks row folds four small classes (5
  circuits): support, buffer, differential pair, and gain stage.\par}
\end{table}

%% file: data/generated/class_status.tex
\begin{table}[t]
  \caption{Verification status by class, computed from the released catalog
  and scoreboard: counted entries, entries holding at least one recorded
  design point, entries validated (some recorded design point passes every
  spec-bounded metric, the T4 criterion of \S\ref{sec:verify}) on at
  least one kit, and entries validated on all three kits. All verdicts are
  typical-corner.}\label{tab:class-status}
  \centering
  \footnotesize
  \setlength{\tabcolsep}{3pt}
  \begin{tabular}{l r r r r}
    \toprule
    & & With & Validated & All 3 \\
    Class & Entries & design pt. & $\geq$1 kit & kits \\
    \midrule
    Amplifiers / OTAs             & 33 & 32 & 15 & 2 \\
    Low-dropout regulators (LDO)  & 7 & 7 & 7 & 6 \\
    Instrumentation amplifiers    & 6 & 5 & 3 & 0 \\
    Support blocks                & 5 & 5 & 3 & 2 \\
    Common-mode feedback (CMFB)   & 4 & 4 & 4 & 0 \\
    Switches                      & 3 & 2 & 2 & 0 \\
    Temperature sensors           & 3 & 3 & 3 & 2 \\
    Comparators                   & 2 & 2 & 2 & 2 \\
    ADC                           & 1 & 1 & 0 & 0 \\
    Line driver                   & 1 & 1 & 1 & 0 \\
    Sampler                       & 1 & 1 & 1 & 0 \\
    Trim                          & 1 & 1 & 1 & 0 \\
    Voltage reference             & 1 & 1 & 1 & 0 \\
    \midrule
    Total & 68 & 65 & 43 & 14 \\
    \bottomrule
  \end{tabular}
\end{table}

%% file: media/F5d-ldo-005-system.tex
\ctikzset{tripoles/mos style/arrows, transistors/scale=1}
\definecolor{netgray}{gray}{0.42}
\tikzset{
  netlbl/.style={text=netgray, font=\scriptsize\itshape, inner sep=1pt},
  blklbl/.style={font=\footnotesize\itshape, inner sep=1pt},
  termlbl/.style={font=\small, inner sep=1pt},
}
\begin{circuitikz}[font=\small, line width=0.7pt]

  \draw[line width=1pt] (1.1,0.6) -- (15.4,0.6) node[right]{$V_{SS}$};

  \node[op amp, noinv input up] (RA) at (2.9,7.2) {};
  \node[font=\small] at (3.05,7.2) {$A_1$};
  \draw (RA.+) -- ++(-1.4,0) node[ocirc]{} node[left]{$V_\mathrm{REF}$};
  \draw (RA.out) -- (4.8,7.2);
  \coordinate (vrefout) at (4.8,7.2);
  \node[netlbl, above=1pt] at (4.35,7.25) {v\_ref\_out};

  \draw (vrefout) -- (4.8,6.6);
  \draw (4.8,6.6) to[R, l=\scriptsize $R_1$] (4.8,4.6);
  \coordinate (vreffb) at (4.8,4.6);
  \draw (4.8,4.6) to[R, l=\scriptsize $R_2$] (4.8,1.9);
  \draw (4.8,1.9) -- (4.8,0.6);
  \node[netlbl, right=1pt] at (4.9,4.55) {v\_ref\_fb};
  \draw (vreffb) -- (1.15,4.6) -- (1.15,6.7) -- (RA.-);

  \draw (vrefout) -- (5.55,7.2);
  \draw (5.55,7.2) to[R, l=\scriptsize $R_\mathrm{lpf}$] (7.2,7.2);
  \coordinate (vlpfout) at (7.2,7.2);
  \node[netlbl, above=1pt] at (7.2,7.35) {v\_lpf\_out};
  \draw (vlpfout) -- (7.2,6.0);
  \draw (7.2,6.0) to[C, l_=\scriptsize $C_\mathrm{lpf}$] (7.2,3.9);
  \draw (7.2,3.9) -- (7.2,0.6);

  \node[op amp] (EA) at (9.7,6.4) {};
  \node[font=\small] at (9.85,6.4) {$A_2$};
  \draw (vlpfout) -- (8.05,7.2) -- (8.05,6.9) -- (EA.-);
  \draw (EA.out) -- (11.85,6.4);
  \coordinate (eaout) at (11.85,6.4);
  \node[netlbl, below=1pt] at (11.2,6.35) {ea\_out};

  \node[pmos] (Mp) at (12.9,7.8) {};
  \node[font=\small] at (13.55,7.9) {$M_\mathrm{p}$};
  \node[blklbl] at (13.75,7.4) {pass};
  \draw (eaout) -- (11.85,7.8) -- (Mp.G);
  \draw (Mp.S) -- (12.9,9.3) node[ocirc]{} node[above]{$V_\mathrm{IN}$};
  \draw (Mp.D) -- (12.9,4.5);
  \coordinate (vout) at (12.9,4.5);

  \draw (vout) -- (15.1,4.5) node[ocirc]{} node[right]{$V_\mathrm{OUT}$};
  \node[netlbl, above=1pt] at (13.5,4.55) {VOUT};
  \draw (13.7,4.5) -- (13.7,4.1);
  \draw (13.7,4.1) to[R, l_=\scriptsize $R_3$] (13.7,1.6);
  \draw (13.7,1.6) -- (13.7,0.6);
  \draw (14.4,4.5) -- (14.4,4.1);
  \draw (14.4,4.1) to[C, l=\scriptsize $C_1$] (14.4,1.6);
  \draw (14.4,1.6) -- (14.4,0.6);
  \draw (vout) -- (12.9,2.6) -- (8.15,2.6) -- (8.15,5.9) -- (EA.+);
  \node[netlbl, above=3pt] at (10.5,2.6) {VOUT feedback (no divider)};

  \node[circ] at (vrefout){};   \node[circ] at (vreffb){};
  \node[circ] at (vlpfout){};   \node[circ] at (vout){};
  \node[circ] at (13.7,4.5){};  \node[circ] at (14.4,4.5){};
  \node[circ] at (4.8,0.6){};   \node[circ] at (7.2,0.6){};
  \node[circ] at (13.7,0.6){};  \node[circ] at (14.4,0.6){};

  \draw[dashed, teal!70!black] (0.2,1.3) rectangle (7.9,8.5);
  \node[font=\footnotesize, text=teal!70!black, anchor=north west] at (0.3,8.45)
        {Buffered reference};

  \draw[dashed, violet] (8.4,3.2) rectangle (15.3,9.7);
  \node[font=\footnotesize, text=violet, anchor=north] at (11.2,9.6)
        {Regulation loop};

\end{circuitikz}

%% file: media/F5e-annotated-ldo-005-error-amp.tex
\ctikzset{tripoles/mos style/arrows, transistors/scale=1}
\definecolor{netgray}{gray}{0.45}
\tikzset{
  netlbl/.style={text=netgray, font=\scriptsize\itshape, inner sep=1pt},
  devlbl/.style={font=\scriptsize, inner sep=1pt},
}
\begin{circuitikz}[font=\small, line width=0.7pt]

  \draw[line width=1pt] (0.5,12.0) -- (16.6,12.0) node[right]{$V_{DD}$};
  \draw[line width=1pt] (0.5,1.0)  -- (16.6,1.0)  node[right]{$V_{SS}$};

  \node[pmos, xscale=-1] (M_bias) at (2.0,10.3) {};        %
  \node[nmos, xscale=-1] (M_mirror_error_amp_ref)  at (2.0,3.2)  {};        %
  \draw (M_bias.S) -- (2.0,12.0);                          %
  \draw (M_mirror_error_amp_ref.S)  -- (2.0,1.0);                           %
  \coordinate (n2_biasgate) at (2.0,9.2);
  \coordinate (n2_refgate)  at (2.0,4.3);
  \draw (M_bias.D) -- (M_mirror_error_amp_ref.D);
  \node[netlbl, left=2pt] at (2.0,6.7) {\#net2};
  \draw (M_bias.G) |- (n2_biasgate);
  \draw (M_mirror_error_amp_ref.G)  |- (n2_refgate);
  \draw (M_mirror_error_amp_ref.G)  -- (2.98,2.5);                          %
  \node[devlbl, left=1pt] at (1.35,10.3) {M\textsubscript{bias}};
  \node[devlbl, left=1pt] at (1.15,3.2)  {M\textsubscript{ref}};

  \draw (2.98,2.5) -- (13.52,2.5);

  \node[pmos, xscale=-1] (M3m) at (8.0,10.3) {};          %
  \node[pmos]            (M5)  at (11.0,10.3){};          %
  \draw (M3m.S) -- (8.0,12.0);
  \draw (M5.S)  -- (11.0,12.0);
  \node[devlbl] at (7.4,10.3) {M3m};
  \node[devlbl] at (11.75,10.3) {M5};

  \draw (5.5,10.95) -- (10.02,10.95);
  \node[netlbl, above=1pt] at (6.2,10.95) {net\_d};
  \draw (M3m.G) -- (8.98,10.95);                          %
  \draw (M5.G)  -- (10.02,10.95);                         %

  \node[pmos, xscale=-1] (M4) at (5.5,8.7) {};            %
  \node[nmos]            (M_mirror_f) at (5.5,3.2) {};            %
  \draw (M4.S) -- (5.5,10.95);                            %
  \draw (M4.D) -- (M_mirror_f.D);                                 %
  \node[netlbl, right=1pt] at (5.5,6.2) {net\_f};
  \draw (M_mirror_f.S) -- (5.5,1.0);
  \draw (M_mirror_f.G) -- (4.52,2.5);                             %
  \node[devlbl] at (4.75,8.7) {M4};
  \node[devlbl, right=1pt] at (6.05,3.2) {M\textsubscript{f}};

  \node[nmos]            (M1) at (8.0,7.2) {};            %
  \node[nmos, xscale=-1] (M2) at (11.0,7.2){};            %
  \node[nmos]            (M_mirror_c) at (9.5,3.2) {};            %
  \draw (M1.G) -- ++(-0.8,0) node[ocirc]{} node[left]{$v_\mathrm{inn}$};
  \draw (M2.G) -- ++(0.8,0)  node[ocirc]{} node[right]{$v_\mathrm{inp}$};
  \node[devlbl] at (8.6,6.75) {M1};
  \node[devlbl] at (10.4,6.75) {M2};
  \node[devlbl] at (10.1,3.2) {M\textsubscript{c}};
  \coordinate (netc) at (9.5,5.5);
  \draw (M1.S) |- (netc);
  \draw (M2.S) |- (netc);
  \draw (M_mirror_c.D) -- (netc);
  \draw (M_mirror_c.S) -- (9.5,1.0);
  \draw (M_mirror_c.G) -- (8.52,2.5);                             %
  \node[netlbl, right=1pt] at (9.6,5.05) {net\_c};

  \coordinate (na_reand) at (8.0,9.3);
  \coordinate (na_m4g)   at (8.0,8.7);
  \draw (M3m.D) -- (M1.D);
  \node[netlbl, right=1pt] at (8.15,8.25) {net\_a};
  \draw (M4.G) -- (na_m4g);                               %

  \coordinate (nb_m6g) at (11.0,9.0);
  \coordinate (nb_rz)  at (11.0,8.3);
  \draw (M5.D) -- (M2.D);
  \node[netlbl, left=1pt] at (10.85,8.25) {net\_b};

  \draw (na_reand) -- (6.6,9.3);                          %
  \draw (6.6,9.3) to[R, l_=\scriptsize $R_\mathrm{EAnd}$] (6.6,10.5);
  \draw (6.6,10.5) -- (6.6,10.95);                        %

  \node[pmos] (M6)  at (14.5,10.3) {};                    %
  \node[nmos] (M_mirror_ea_out) at (14.5,3.2)  {};                    %
  \draw (M6.S)  -- (14.5,12.0);
  \draw (M_mirror_ea_out.S) -- (14.5,1.0);
  \draw (M_mirror_ea_out.G) -- (13.52,2.5);                           %
  \node[devlbl] at (15.15,10.3) {M6};
  \node[devlbl] at (15.35,3.2)  {M\textsubscript{eo}};
  \coordinate (vout_cc)  at (14.5,8.3);
  \coordinate (vout_pin) at (14.5,6.5);
  \draw (M6.D) -- (M_mirror_ea_out.D);
  \node[netlbl, right=1pt] at (14.6,4.6) {vout};
  \draw (M6.G) -- (13.52,9.0) -- (nb_m6g);
  \draw (vout_pin) -- (16.2,6.5) node[ocirc]{} node[right]{$v_\mathrm{out}$};

  \draw (nb_rz) -- (11.35,8.3);
  \draw (11.35,8.3) to[R, l=\scriptsize $R_Z$] (12.7,8.3);
  \coordinate (net1) at (12.7,8.3);
  \draw (net1) to[C, l=\scriptsize $C_c$] (13.95,8.3);
  \draw (13.95,8.3) -- (vout_cc);
  \node[netlbl, above=0pt] at (12.7,8.55) {\#net1};

  \node[circ] at (2.0,12){};   \node[circ] at (8.0,12){};
  \node[circ] at (11.0,12){};  \node[circ] at (14.5,12){};
  \node[circ] at (2.0,1){};    \node[circ] at (5.5,1){};
  \node[circ] at (9.5,1){};    \node[circ] at (14.5,1){};
  \node[circ] at (n2_biasgate){};   \node[circ] at (n2_refgate){};
  \node[circ] at (2.98,3.2){};      %
  \node[circ] at (4.52,2.5){};      %
  \node[circ] at (8.52,2.5){};      %
  \node[circ] at (na_reand){};      \node[circ] at (na_m4g){};
  \node[circ] at (nb_m6g){};        \node[circ] at (nb_rz){};
  \node[circ] at (netc){};
  \node[circ] at (6.6,10.95){};     \node[circ] at (8.98,10.95){};
  \node[circ] at (vout_cc){};       \node[circ] at (vout_pin){};

  \draw[dashed, teal] (1.15,1.7) rectangle (2.9,11.2);
  \node[font=\scriptsize, text=teal, anchor=south] at (2.0,11.25) {Self-bias ref.};

  \draw[dashed, blue] (6.7,6.05) rectangle (11.9,7.7);
  \node[font=\scriptsize, text=blue, anchor=north] at (9.5,6.0) {Input pair};

  \draw[dashed, orange!85!black] (4.5,7.8) rectangle (11.8,11.4);
  \node[font=\scriptsize, text=orange!85!black, anchor=south west] at (4.6,11.45) {Level-shifted mirror load};

  \draw[dashed, teal!80!black] (4.3,1.75) rectangle (10.5,3.9);
  \node[font=\scriptsize, text=teal!80!black, anchor=north west] at (4.3,1.7) {NMOS tail + sinks};

  \draw[dashed, violet] (13.6,1.8) rectangle (15.5,11.2);
  \node[font=\scriptsize, text=violet, anchor=south] at (14.55,11.25) {Output stage};

\end{circuitikz}

%% file: data/generated/ldo_spec_bands.tex
\begin{table*}[t]
  \caption{Per-entry datasheet specification bands and operating conditions
  for the LDO benchmark (Table~\ref{tab:ldo-ppa}, whose row order and
  accession numbers these share), read mechanically from each entry's
  released datasheet and bench bindings. The last header row gives each
  column's bound direction.}\label{tab:ldo-specs}
  \centering
  \footnotesize
  \setlength{\tabcolsep}{5pt}
  \begin{tabular}{@{}l c c c c c c c c c c c c c@{}}
    \toprule
    & \multicolumn{5}{c}{Conditions}
    & \multicolumn{8}{c}{Specification band} \\
    \cmidrule(lr){2-6} \cmidrule(lr){7-14}
    Circuit & $V_{\mathrm{in}}$ & $C_{\mathrm{out}}$ &
    $I_{\mathrm{load}}$ & Load step & Load range & $V_{\mathrm{out}}$ &
    Dropout & Load reg. & Line reg. & PSRR &
    $Z_{\mathrm{out}}$ & Under. & $I_q$ \\
    & (V) & & (mA) & (mA) & (mA) & (V) & (mV) & (mV) & (mV) & (dB) & (dB) & (mV) &
    ($\mu$A) \\
    & & & & & & window & max & max & max & min & max & max & max \\
    \midrule
    \texttt{001} & 1.8 & 50\,pF & 5 & 5$\to$55 & 0--55 & 1.50--1.70 & 300 & 800 & 300 & 5 & 35 & 1500 & 100 \\
    \texttt{002} & 2 & 50\,pF & 10 & 0.01$\to$10 & 0--10 & 1.70--1.90 & 300 & 20 & 50 & 40 & 45 & 1000 & 500 \\
    \texttt{003} & 2 & 50\,pF & 10 & 0.01$\to$10 & 0--10 & 1.70--1.95 & 300 & 150 & 50 & 20 & 25 & 1200 & 500 \\
    \texttt{004} & 1.8 & 1\,$\mu$F & 1 & 1$\to$10 & 0--10 & 1.25--1.35 & 300 & 150 & 30 & 20 & 15 & 150 & 250 \\
    \texttt{005} & 3.3 & 1\,$\mu$F & 1 & 1$\to$10 & 0--10 & 1.50--1.70 & 400 & 60 & 300 & 25 & 10 & 150 & 800 \\
    \texttt{007} & 1.8 & 1\,$\mu$F & 1 & 1$\to$10 & 0--10 & 1.16--1.24 & 300 & 20 & 20 & 30 & 8 & 100 & 250 \\
    \texttt{008} & 1.8 & 1\,$\mu$F & 1 & 1$\to$10 & 0--10 & 1.10--1.30 & 500 & 100 & 50 & 20 & 20 & 300 & 500 \\
    \texttt{009} & 1.8 & 1\,$\mu$F & 1 & 1$\to$10 & 0--10 & 0.85--0.95 & 500 & 100 & 50 & 20 & 20 & 300 & 100 \\
    \bottomrule
  \end{tabular}
  \vspace{2pt}

  {\footnotesize Bands are authored per entry, so ``passes its datasheet'' is a per-entry claim: the relaxed \texttt{001} bands (a capacitor-less regulator) and the tight \texttt{005} bands are both visible rather than hidden behind one verdict.\par}

  {\footnotesize The dropout bench of \texttt{007} at 10\,mA against a 1\,mA nominal load runs heavier than nominal; every other entry's dropout is judged at its nominal load.\par}
\end{table*}

%% file: data/generated/amp_native_loads.tex
\begin{table}[h]
  \caption{Cells evaluated 10$\times$ or more from their as-published load under the
  uniform 10\,pF condition. The uniform load is what makes the cross-corpus comparison a
  comparison; the native values are kept so each result can also be read against its
  source publication's operating point.}
  \label{tab:amp-native-loads}
  \centering\small
  \begin{tabular}{@{}ll@{}}
    \toprule Circuit & Native $C_L$ \\ \midrule
    amp\_021 & 15\,nF \\
    amp\_013 & 1.5\,nF \\
    amp\_017 & 560\,pF \\
    amp\_015 & 500\,pF \\
    amp\_016 & 500\,pF \\
    amp\_011 & 150\,pF \\
    amp\_012 & 150\,pF \\
    amp\_002 & 100\,pF \\
    amp\_005 & 100\,pF \\
    amp\_006 & 100\,pF \\
    amp\_007 & 100\,pF \\
    amp\_010 & 100\,pF \\
    amp\_014 & 100\,pF \\
    amp\_022 & 500\,fF \\
    amp\_023 & 500\,fF \\
    amp\_001 & 50\,fF \\
    amp\_018 & 50\,fF \\
    amp\_025 & 50\,fF \\
    amp\_026 & 50\,fF \\
    amp\_027 & 50\,fF \\
    amp\_029 & 50\,fF \\
    amp\_030 & 50\,fF \\
    amp\_031 & 50\,fF \\
    \bottomrule
  \end{tabular}
\end{table}

%% file: data/generated/amp_ppa_ihp-sg13g2.tex
\begin{table}[p]
  \caption{Amplifier/OTA benchmark in IHP 130\,nm (\texttt{ihp-sg13g2}), at
  the released sizing of each entry. All values are re-simulated under uniform conditions ($C_L$=10\,pF, unity-gain buffer; noise 1\,Hz--1\,MHz; THD per the footnote).}\label{tab:amp-ppa}
  \centering
  \footnotesize
  \setlength{\tabcolsep}{2pt}
  \resizebox{\linewidth}{!}{%
  \begin{tabular}{l c l r r r r r r r r r r}
    \toprule
    Circuit & Stages & Comp. & Gain (dB) & UGF (MHz) & PM ($^\circ$) & CMRR (dB) & PSRR (dB) & Power ($\mu$W) & Gate area ($\mu$m$^2$) & $v_{n,\mathrm{in}}$ & $v_{n,\mathrm{in}}^{\mathrm{BW}}$ & THD \\
    \midrule
    \texttt{001\_5t} & 1 & none & 32.5 & 1.21 & 86.9 & 49.5 & 32.7 & 59.3 & 348 & 30.9 & 33.2 & 0.109 \\
    \texttt{002\_alfio\_raffc}$^{\dagger}$\(^{\S}\) & 3 & RAFFC & 103.9 & 2.41 & 75.1 & 82.8 & 83.4 & 605.2 & 1892 & 40.9 & 66.6 & 0.012 \\
    \texttt{003\_fan\_smc}$^{\dagger}$ & 3 & SMC & 86.0 & 6.64 & 85.8 & 67.8 & 54.4 & 1589.4 & 412 & 189.0 & 219.5 & 0.020 \\
    \texttt{004\_folded\_cascode} & 1 & none & 44.1 & 2.03 & 74.8 & 72.9 & 39.6 & 233.0 & 3556 & 42.0 & 67.7 & 0.769 \\
    \texttt{005\_hoilee\_affc}$^{\dagger}$\(^{\S}\) & 3 & AFFC & 97.5 & 3.50 & 103.8 & 74.9 & 80.5 & 342.0 & 2200 & 39.7 & 76.3 & 0.026 \\
    \texttt{006\_leung\_dfcfc1}$^{\dagger}$\(^{\S}\) & 3 & DFCFC1 & 93.8 & 1.92 & 75.4 & 87.4 & 57.7 & 692.3 & 12803 & 17.7 & 24.4 & 0.001 \\
    \texttt{007\_leung\_dfcfc2}$^{\dagger}$ & 3 & DFCFC2 & 91.3 & 2.78 & 78.4 & 90.6 & 63.5 & 448.7 & 440 & 52.9 & 80.6 & 0.001 \\
    \texttt{008\_leung\_nmcf}$^{\dagger}$\(^{\S}\) & 3 & NMCF & 89.4 & 3.16 & 62.7 & 86.6 & 65.6 & 315.2 & 646 & 65.1 & 86.4 & 0.003 \\
    \texttt{009\_leung\_nmcnr}$^{\dagger}$ & 3 & NMCNR & 95.6 & 1.44 & 89.3 & 91.0 & 63.2 & 117.6 & 272 & 106.2 & 115.7 & 0.001 \\
    \texttt{010\_peng\_acbc}$^{\dagger}$\(^{\S}\) & 3 & ACBC & 84.4 & 1.31 & 86.4 & 86.8 & 49.8 & 648.5 & 1549 & 83.2 & 87.2 & 0.003 \\
    \texttt{011\_peng\_iac}$^{\dagger}$\(^{\S}\) & 3 & IAC & 109.6 & 1.76 & 82.3 & 133.6 & 64.5 & 37.1 & 1273 & 89.2 & 127.9 & -- \\
    \texttt{012\_peng\_tcfc}$^{\dagger}$\(^{\S}\) & 3 & TCFC & 102.3 & 2.01 & 71.4 & 100.9 & 63.9 & 122.4 & 1691 & 93.8 & 122.9 & -- \\
    \texttt{013\_qu2017\_azc}$^{\dagger}$ & 3 & AZC & 80.1 & 3.78 & 88.7 & 90.2 & 76.8 & 87.4 & 204 & 291.3 & 383.0 & -- \\
    \texttt{014\_ramos\_pfc}$^{\dagger}$ & 3 & PFC & 102.1 & 1.94 & 81.5 & 102.6 & 61.8 & 754.5 & 2583 & 20.6 & 27.2 & $<$0.001 \\
    \texttt{015\_sau\_cfcc}$^{\dagger}$ & 3 & CFCC & 87.4 & 7.06 & 82.5 & 77.6 & 56.5 & 295.2 & 4351 & 42.3 & 82.4 & 1.069 \\
    \texttt{016\_song\_dacfc}$^{\dagger}$ & 3 & DACFC & 72.1 & 1.91 & 106.4 & 75.1 & 39.7 & 252.6 & 472 & 395.4 & 475.5 & 0.165 \\
    \texttt{018\_telescopic\_cascode}\(^{\S}\) & 1 & none & 52.2 & 0.30 & 88.9 & 96.5 & 52.7 & 11.4 & 97 & 125.3 & 113.2 & 0.017 \\
    \texttt{019\_ti\_ldo\_error}\(^{\S}\) & 2 & none & 50.0 & 11.63 & 74.3 & 54.8 & 51.2 & 10505.3 & 437 & 20.3 & 31.2 & 0.020 \\
    \texttt{021\_yan\_az}$^{\dagger}$\(^{\S}\) & 3 & AZ & 129.4 & 2.19 & 92.0 & 43.1 & 51.9 & 121.0 & 1258 & 72.4 & 80.6 & 2.094 \\
    \texttt{022\_fer\_two\_stage} & 2 & miller & 51.5 & 19.16 & 63.4 & 64.3 & 76.4 & 369.0 & 176 & 44.4 & 82.2 & 0.022 \\
    \texttt{023\_fer\_fd2s} & 2 & miller\_nulling & 65.9 & 25.25 & 73.6 & 95.8 & 100.2 & 824.6 & 92 & 79.4 & 116.3 & 0.001 \\
    \texttt{024\_smcnr}$^{\dagger}$ & 2 & miller\_nulling & 69.4 & 0.15 & 58.2 & 96.4 & 105.5 & 2.0 & 332 & 280.6 & 71.4 & 0.015 \\
    \texttt{025\_hsu\_classab\_ota}\(^{\S}\) & 2 & miller-rz & 35.8 & 49.14 & 53.5 & 98.9 & 127.7 & 208.3 & 123 & 1130.9 & 6613.9 & 6.589 \\
    \texttt{026\_fan\_chopper\_ota} & 2 & miller & 59.7 & 6.16 & 59.4 & 10.4$^{\P}$ & -2.4$^{\P}$ & 268.9 & 141 & 76.1 & 95.8 & $<$0.001 \\
    \texttt{027\_fan\_rrl\_ota} & 1 & \textemdash & 46.6 & 3.61 & 89.9 & 52.8 & 63.4 & 77.1 & 83 & 56.3 & 65.2 & 0.006 \\
    \texttt{030\_miller\_cmfb\_composite} & 2 & miller & 50.1 & 13.76 & 33.3 & 80.6 & 50.2 & 218.0 & 58 & 42.2 & 73.8 & 0.002 \\
    \texttt{031\_srmc\_core\_cmfb}~\cite{analogdb_design_SRMC_iahsu68DBTHD2020} & 2 & miller-rz & 54.7 & 5.84 & 71.3 & 145.1 & 114.6 & 116.2 & 14 & 59.4 & 72.7 & -- \\
    \texttt{017\_tan\_clia}$^{\dagger}$ & 3 & CLIA & 43.8 & 1.95 & 59.9 & 49.8 & 56.1 & 123.3 & 11128 & 56.1 & 74.5 & 0.380 \\
    \texttt{032\_ti\_ldo\_error\_selfbias} & 2 & Rz+Cc$^{\ast}$ & 50.0 & 18.28 & 69.6 & 56.5 & 57.0 & 195.2 & 1443 & 40.7 & 73.6 & 0.075 \\
    \texttt{033\_ti\_ldo\_ref\_selfbias} & 2 & Rz+Cc$^{\ast}$ & 56.1 & 29.17 & 80.3 & 50.7 & 68.7 & 236.9 & 37 & 52.3 & 82.1 & 0.008 \\
    \texttt{034\_fan\_chopper\_cmfb} & 2 & miller & 58.9 & 6.11 & 58.2 & 53.8 & 41.0 & 272.2 & 148 & 76.1 & -- & -- \\
    \texttt{035\_fan\_chopper\_cmfb\_dual} & 2 & miller & 60.3 & 6.20 & 62.9 & 110.9 & 90.4 & 328.0 & 156 & 76.2 & -- & -- \\
    \bottomrule
  \end{tabular}}
  \vspace{2pt}

  {\footnotesize \textemdash{} and -- mark a cell with no value: not
  applicable, not recorded in this configuration, or a recorded bench
  failure. $^{\dagger}$AnalogGym-sourced, ported from its native sky130
  testbenches.\par}

  {\footnotesize \(^{\S}\)Re-sized after failing stability verification
  (imported phase margin below \(45^\circ\)); re-optimization constrained
  it to 60--90\(^\circ\). \texttt{002} was conditionally unstable and was
  re-sized to a 5.7\,dB gain margin.\par}

  {\footnotesize $^{\ast}$Abbreviated compensation scheme: Rz+Cc = ``Rz + Cc from the first-stage node to the output''.\par}

  {\footnotesize Fully-differential rows report the systematic
  CMRR/PSRR (mismatch excluded), measured on the common-mode path
  with the differential gain folded in.\par}

  {\footnotesize $^{\P}$Chopper core characterized with its output chopper held static; chopping leaves the CM path unchanged, and with no CMFB the rejection is CMFB-limited. Adding one (\texttt{034\_fan\_chopper\_cmfb}) improves both CM-path columns by 43.4\,dB on ihp-sg13g2 for 0.8\,dB of differential gain.\par}

  {\footnotesize PM is the stb loop-probe reading (\texttt{pm\_loop\_deg}) where that bench's unity crossing agrees with the AC sweep to within 20\,\%, and the AC-sweep reading (\texttt{pm\_deg}) otherwise; either is folded into $(-180^\circ,180^\circ]$. Here every tabulated PM cell is the stb loop-probe reading. Both readings are carried per design point in the released scoreboard.\par}

  {\footnotesize Both noise columns are $\mu$V$_{rms}$; THD is \%
  (single-tone, 0.1\,V, or 0.05\,V where the swing would clip, at a
  per-cell frequency set from its UGF, harmonics 2--5).\par}

  {\footnotesize Gate area is the summed gate area of the sized devices only; capacitor and resistor area is excluded (\S\ref{sec:scoreboard}).\par}

  {\footnotesize \texttt{017} is tabulated for information but excluded from the counted corpus; the counted \texttt{028} (behavioral) and \texttt{029} (no recorded baseline) are not tabulated.\par}

  {\footnotesize The $\sim$10\,mW of \texttt{019\_ti\_ldo\_error} is design intent: the source design's 1\,mA reference is mirrored at near-unity ratio into the tail and output legs.\par}

  {\footnotesize $v_{n,\mathrm{in}}$ integrates the uniform 1\,Hz--1\,MHz
  comparison band, $v_{n,\mathrm{in}}^{\mathrm{BW}}$ each cell's own
  1\,Hz--UGF band; the uniform band overstates a slow cell (up to
  7.8$\times$) and understates a fast one (down to 0.17$\times$).\par}
\end{table}

%% file: data/generated/amp_ppa_gf180mcu.tex
\begin{table}[p]
  \caption{Amplifier/OTA benchmark in GF180MCU (\texttt{gf180mcu}): every
  amplifier bound to this kit with a recorded AC baseline, at its
  released sizing. All values are re-simulated under uniform conditions ($C_L$=10\,pF, unity-gain buffer; noise 1\,Hz--1\,MHz; THD per the footnote).}\label{tab:amp-ppa-gf}
  \centering
  \footnotesize
  \setlength{\tabcolsep}{2pt}
  \resizebox{\linewidth}{!}{%
  \begin{tabular}{l c l r r r r r r r r r r}
    \toprule
    Circuit & Stages & Comp. & Gain (dB) & UGF (MHz) & PM ($^\circ$) & CMRR (dB) & PSRR (dB) & Power ($\mu$W) & Gate area ($\mu$m$^2$) & $v_{n,\mathrm{in}}$ & $v_{n,\mathrm{in}}^{\mathrm{BW}}$ & THD \\
    \midrule
    \texttt{001\_5t}\(^{\S}\) & 1 & none & 27.9 & 0.02 & 90.1 & 51.0 & 53.5 & 30.5 & 40 & 206.2 & 57.7 & 8.743 \\
    \texttt{002\_alfio\_raffc}$^{\dagger}$\(^{\S}\) & 3 & RAFFC & 90.1 & 3.75 & 84.4 & 89.2 & 63.4 & 422.5 & 8419 & 25.1 & 55.0 & 0.005 \\
    \texttt{003\_fan\_smc}$^{\dagger}$\(^{\S}\) & 3 & SMC & 191.7 & 1.02 & 56.9 & 95.6 & 59.4 & 2102.8 & 107196$^{\#}$ & 12.7 & 12.7 & 0.153 \\
    \texttt{004\_folded\_cascode}\(^{\S}\) & 1 & none & 77.6 & 10.77 & 87.1 & 121.7 & 83.5 & 998.4 & 918 & 23.9 & 42.7 & $<$0.001 \\
    \texttt{005\_hoilee\_affc}$^{\dagger}$\(^{\S}\) & 3 & AFFC & 64.0 & 1.88 & 65.1 & 131.3 & 75.3 & 857.2 & 147597$^{\#}$ & 27.2 & 38.5 & 0.128 \\
    \texttt{006\_leung\_dfcfc1}$^{\dagger}$\(^{\S}\) & 3 & DFCFC1 & 163.6 & 1.32 & 66.1 & 96.6 & 60.6 & 1905.3 & 164833$^{\#}$ & 27.0 & 36.3 & 0.001 \\
    \texttt{007\_leung\_dfcfc2}$^{\dagger}$\(^{\S}\) & 3 & DFCFC2 & 152.5 & 1.48 & 63.7 & 116.6 & 64.1 & 729.4 & 31709 & 35.5 & 52.3 & $<$0.001 \\
    \texttt{008\_leung\_nmcf}$^{\dagger}$\(^{\S}\) & 3 & NMCF & 172.4 & 0.77 & 62.9 & 110.6 & 57.6 & 700.9 & 47396 & 25.1 & 20.4 & 0.033 \\
    \texttt{009\_leung\_nmcnr}$^{\dagger}$\(^{\S}\) & 3 & NMCNR & 159.6 & 2.44 & 67.7 & 97.8 & 70.3 & 588.1 & 24595 & 22.3 & 63.0 & $<$0.001 \\
    \texttt{010\_peng\_acbc}$^{\dagger}$\(^{\S}\) & 3 & ACBC & 162.5 & 0.51 & 70.5 & 116.6 & 53.3 & 191.6 & 22939 & 46.4 & 28.6 & $<$0.001 \\
    \texttt{012\_peng\_tcfc}$^{\dagger}$\(^{\S}\) & 3 & TCFC & 155.2 & 1.12 & 66.5 & 85.9 & 60.5 & 323.7 & 24526 & 41.4 & 43.7 & 0.002 \\
    \texttt{013\_qu2017\_azc}$^{\dagger}$\(^{\S}\) & 3 & AZC & 121.8 & 2.61 & 65.4 & 70.9 & 70.5 & 264.6 & 21107 & 50.5 & 92.2 & 0.680 \\
    \texttt{014\_ramos\_pfc}$^{\dagger}$\(^{\S}\) & 3 & PFC & 184.1 & 0.50 & 67.4 & 110.2 & 52.3 & 2294.4 & 277287$^{\#}$ & 10.7 & 5.8 & 0.001 \\
    \texttt{015\_sau\_cfcc}$^{\dagger}$\(^{\S}\) & 3 & CFCC & 150.9 & 6.76 & 95.5 & 100.1 & 85.3 & 1688.8 & 36759 & 23.5 & 64.2 & 4.082 \\
    \texttt{016\_song\_dacfc}$^{\dagger}$ & 3 & DACFC & 63.4 & 7.71 & 81.0 & 48.7 & 57.5 & 587.3 & 10302 & 149.8 & 240.2 & 0.037 \\
    \texttt{017\_tan\_clia}$^{\dagger}$\(^{\S}\) & 3 & CLIA & 75.4 & 2.58 & 51.7 & 124.0 & 86.4 & 166.5 & 27722 & 117.4 & 218.1 & 0.093 \\
    \texttt{018\_telescopic\_cascode}\(^{\S}\) & 1 & none & 82.7 & 0.48 & 89.9 & 95.0 & 82.3 & 11.5 & 71 & 49.8 & 38.5 & 0.029 \\
    \texttt{019\_ti\_ldo\_error}\(^{\S}\) & 2 & none & 43.0 & 6.32 & 91.4 & 42.0 & 67.7 & 7497.1 & 302 & 31.5 & 38.0 & 0.429 \\
    \texttt{022\_fer\_two\_stage} & 2 & miller & 69.7 & 9.54 & 66.5 & 49.3 & 49.5 & 226.4 & 176 & 46.0 & 80.0 & 1.767 \\
    \texttt{023\_fer\_fd2s}\(^{\S}\) & 2 & miller\_nulling & 78.9 & 16.05 & 62.6 & 102.9 & 108.9 & 584.8 & 95 & 82.1 & 112.0 & 0.004 \\
    \texttt{024\_smcnr}$^{\dagger}$\(^{\S}\) & 2 & miller\_nulling & 116.0 & 0.08 & 65.7 & 72.0 & 71.6 & 1.2 & 332 & 581.2 & 74.6 & 1.544 \\
    \texttt{032\_ti\_ldo\_error\_selfbias}\(^{\S}\) & 2 & none & 72.7 & 18.65 & 82.7 & 61.7 & 56.0 & 590.0 & 460 & 19.4 & 48.0 & 0.011 \\
    \texttt{033\_ti\_ldo\_ref\_selfbias} & 2 & none & 74.1 & 7.29 & 74.1 & 43.2 & 54.1 & 390.5 & 331 & 18.0 & 38.7 & 0.304 \\
    \texttt{025\_hsu\_classab\_ota} & 2 & miller-rz & 39.9 & 0.36 & 124.7 & 118.0 & 75.4 & 1752.1 & 3406 & 28.2 & 21.2 & -- \\
    \texttt{026\_fan\_chopper\_ota} & 2 & miller & 27.1 & 0.63 & 75.0 & 62.6$^{\P}$ & -0.9$^{\P}$ & 920.9 & 2369 & 36.7 & 29.8 & 3.270 \\
    \texttt{027\_fan\_rrl\_ota} & 1 & None & 71.4 & 1.36 & 87.9 & 132.1 & 66.3 & 463.0 & 1317 & 46.1 & 49.8 & $<$0.001 \\
    \texttt{031\_srmc\_core\_cmfb} & 2 & miller-rz & 65.9 & 7.80 & 68.9 & $>$180 & 125.4 & 985.0 & 14 & 65.2 & 82.5 & -- \\
    \bottomrule
  \end{tabular}}
  \vspace{2pt}

  {\footnotesize \textemdash{} and -- mark a cell with no value: not
  applicable, not recorded in this configuration, or a recorded bench
  failure. $^{\dagger}$AnalogGym-sourced. \(^{\S}\)Re-sized after failing cross-kit verification. \texttt{004\_folded\_cascode} retains its recorded
  values, but its deck does not simulate on this kit at the archived
  revision (an unresolved device-model defect), so they are not
  reproducible here; \texttt{017\_tan\_clia} was re-sized on this kit and
  its released point is stable ($+51.7^\circ$).\par}

  {\footnotesize Rejection readings at or above 180\,dB sit at the AC solver's resolution floor and are printed as bounds; every CMRR/PSRR figure is systematic (mismatch excluded), an upper bound (\S\ref{sec:datasheet}).\par}

  {\footnotesize $^{\#}$Constrained re-optimizer output, stability-feasible only above 0.1\,mm$^2$ of summed gate area, an optimizer artifact rather than a defensible design; no marked row passes its full datasheet, so none enters the validated set of Fig.~\ref{fig:ppa-landscape}.\par}

  {\footnotesize Fully-differential rows report the systematic
  CMRR/PSRR (mismatch excluded), measured on the common-mode path
  with the differential gain folded in.\par}

  {\footnotesize $^{\P}$Chopper core characterized with its output chopper held static; chopping leaves the CM path unchanged, and with no CMFB the rejection is CMFB-limited. Adding one (\texttt{034\_fan\_chopper\_cmfb}) improves both CM-path columns by 43.4\,dB on ihp-sg13g2 for 0.8\,dB of differential gain.\par}

  {\footnotesize PM is the stb loop-probe reading (\texttt{pm\_loop\_deg}) where that bench's unity crossing agrees with the AC sweep to within 20\,\%, and the AC-sweep reading (\texttt{pm\_deg}) otherwise; either is folded into $(-180^\circ,180^\circ]$. Here 26 of the 27 tabulated PM cells are the stb loop-probe reading and the rest are the AC sweep. Both readings are carried per design point in the released scoreboard.\par}

  {\footnotesize Both noise columns are $\mu$V$_{rms}$; THD is \%
  (single-tone, 0.1\,V, or 0.05\,V where the swing would clip, at a
  per-cell frequency set from its UGF, harmonics 2--5).\par}

  {\footnotesize Gate area is the summed gate area of the sized devices only; capacitor and resistor area is excluded (\S\ref{sec:scoreboard}).\par}

  {\footnotesize \texttt{017} is tabulated for information but excluded from the counted corpus; the counted \texttt{028} (behavioral) and \texttt{029} (no recorded baseline) are not tabulated.\par}

  {\footnotesize The $\sim$10\,mW of \texttt{019\_ti\_ldo\_error} is design intent: the source design's 1\,mA reference is mirrored at near-unity ratio into the tail and output legs.\par}

  {\footnotesize \texttt{031\_srmc\_core\_cmfb} carries the same gate area on ihp-sg13g2 and gf180mcu because the gf180mcu binding ports the ihp geometry unchanged.\par}

  {\footnotesize $v_{n,\mathrm{in}}$ integrates the uniform 1\,Hz--1\,MHz
  comparison band, $v_{n,\mathrm{in}}^{\mathrm{BW}}$ each cell's own
  1\,Hz--UGF band; the uniform band overstates a slow cell (up to
  7.8$\times$) and understates a fast one (down to 0.17$\times$).\par}

  {\footnotesize Withheld entries: for each, no sizing was found that is both stable and within specification, so no design point is claimed. \texttt{011\_peng\_iac} -- the only stabilizing sizing found sacrifices 37\,dB of PSRR and violates its specification; \texttt{021\_yan\_az} -- self-oscillates at 1.84\,V$_{pp}$ with the input held at DC; the oscillation arises inside the cell, where no outer compensation reaches it.\par}
\end{table}

%% file: data/generated/amp_ppa_sky130.tex
\begin{table}[p]
  \caption{Amplifier/OTA benchmark in SKY130 (\texttt{sky130}): every
  amplifier bound to this kit with a recorded AC baseline, at its
  released sizing. All values are re-simulated under uniform conditions ($C_L$=10\,pF, unity-gain buffer; noise 1\,Hz--1\,MHz; THD per the footnote).}\label{tab:amp-ppa-sky}
  \centering
  \footnotesize
  \setlength{\tabcolsep}{2pt}
  \resizebox{\linewidth}{!}{%
  \begin{tabular}{l c l r r r r r r r r r r}
    \toprule
    Circuit & Stages & Comp. & Gain (dB) & UGF (MHz) & PM ($^\circ$) & CMRR (dB) & PSRR (dB) & Power ($\mu$W) & Gate area ($\mu$m$^2$) & $v_{n,\mathrm{in}}$ & $v_{n,\mathrm{in}}^{\mathrm{BW}}$ & THD \\
    \midrule
    \texttt{001\_5t} & 1 & none & 26.0 & 0.30 & 90.5 & 21.7 & 26.8 & 36.1 & 24 & 147.7 & 122.2 & 7.511 \\
    \texttt{002\_alfio\_raffc}$^{\dagger}$\(^{\S}\) & 3 & RAFFC & 109.6 & 2.06 & 68.5 & 80.1 & 78.2 & 677.4 & 28122 & 57.9 & 89.9 & 0.642 \\
    \texttt{003\_fan\_smc}$^{\dagger}$ & 3 & SMC & 58.9 & 2.72 & 88.8 & 36.5 & 20.6 & 1473.9 & 412 & 257.5 & 300.3 & 0.272 \\
    \texttt{004\_folded\_cascode} & 1 & none & 59.3 & 1.77 & 90.4 & 48.2 & 43.8 & 177.7 & 1300 & 99.1 & 114.2 & 0.472 \\
    \texttt{005\_hoilee\_affc}$^{\dagger}$\(^{\S}\) & 3 & AFFC & 91.9 & 2.56 & 109.8 & 50.7 & 43.0 & 209.2 & 315 & 182.8 & 219.8 & 0.117 \\
    \texttt{006\_leung\_dfcfc1}$^{\dagger}$\(^{\S}\) & 3 & DFCFC1 & 122.9 & 2.18 & 87.4 & 80.8 & 65.8 & 488.5 & 2139 & 61.0 & 81.6 & 0.001 \\
    \texttt{007\_leung\_dfcfc2}$^{\dagger}$ & 3 & DFCFC2 & 102.4 & 1.79 & 86.0 & 80.5 & 59.3 & 429.3 & 440 & 101.5 & 143.4 & 0.003 \\
    \texttt{008\_leung\_nmcf}$^{\dagger}$\(^{\S}\) & 3 & NMCF & 116.0 & 2.97 & 65.5 & 44.5 & 44.1 & 326.1 & 1128 & 99.2 & 145.9 & 0.122 \\
    \texttt{009\_leung\_nmcnr}$^{\dagger}$ & 3 & NMCNR & 114.6 & 1.22 & 82.2 & 100.1 & 61.9 & 111.4 & 272 & 119.1 & 128.3 & 0.001 \\
    \texttt{010\_peng\_acbc}$^{\dagger}$ & 3 & ACBC & 90.7 & 0.91 & 88.2 & 43.7 & 42.7 & 424.5 & 240 & 215.3 & 209.4 & 0.118 \\
    \texttt{011\_peng\_iac}$^{\dagger}$\(^{\S}\) & 3 & IAC & 129.5 & 4.07 & 80.5 & 98.1 & 71.2 & 91.8 & 835 & 125.0 & 190.0 & 0.001 \\
    \texttt{012\_peng\_tcfc}$^{\dagger}$\(^{\S}\) & 3 & TCFC & 106.8 & 2.17 & 81.7 & 93.0 & 65.9 & 213.0 & 1210 & 104.3 & 137.1 & 0.001 \\
    \texttt{013\_qu2017\_azc}$^{\dagger}$ & 3 & AZC & 94.2 & 2.90 & 88.6 & 80.8 & 28.4 & 76.1 & 204 & 272.1 & 376.4 & 1.338 \\
    \texttt{014\_ramos\_pfc}$^{\dagger}$ & 3 & PFC & 135.9 & 1.46 & 83.9 & 111.9 & 59.1 & 685.6 & 2583 & 33.1 & 39.6 & 0.002 \\
    \texttt{015\_sau\_cfcc}$^{\dagger}$\(^{\S}\) & 3 & CFCC & \textemdash & \textemdash & \textemdash & 63.6 & 50.4 & 255.4 & 254 & 132.2 & -- & 0.586 \\
    \texttt{016\_song\_dacfc}$^{\dagger}$ & 3 & DACFC & 84.0 & 1.22 & 103.3 & 64.0 & 39.1 & 225.2 & 472 & 583.2 & 612.6 & 0.355 \\
    \texttt{019\_ti\_ldo\_error}\(^{\S}\) & 2 & none & 41.1 & 68.10 & 58.0 & 45.5 & 47.9 & 10267.6 & 524 & 42.9 & 86.0 & 0.128 \\
    \texttt{021\_yan\_az}$^{\dagger}$\(^{\S}\) & 3 & AZ & 153.7 & 5.17 & 119.2 & 59.6 & 61.6 & 126.1 & 144 & 182.1 & 226.5 & 3.441 \\
    \texttt{022\_fer\_two\_stage}\(^{\S}\) & 2 & miller & 66.9 & 10.28 & 68.1 & 69.2 & 74.5 & 250.4 & 176 & 91.6 & 138.6 & 0.162 \\
    \texttt{023\_fer\_fd2s} & 2 & miller\_nulling & 81.3 & 21.16 & 67.0 & 107.3 & 112.6 & 557.2 & 92 & 102.3 & 158.4 & 0.003 \\
    \texttt{024\_smcnr}$^{\dagger}$\(^{\S}\) & 2 & miller\_nulling & 93.3 & 0.12 & 67.4 & 68.9 & 69.0 & 2.3 & 355 & 374.7 & 93.5 & 0.412 \\
    \texttt{017\_tan\_clia}$^{\dagger}$ & 3 & CLIA & 118.3 & 1.97 & 78.5 & 81.5 & 60.4 & 29.0 & 7669 & 280.9 & 337.9 & 0.082 \\
    \texttt{018\_telescopic\_cascode} & 1 & none & 57.5 & 1.80 & 87.7 & 61.3 & 57.5 & 21.7 & 360 & 40.3 & 46.2 & 0.028 \\
    \texttt{025\_hsu\_classab\_ota} & 2 & miller-rz & 47.5 & 0.35 & 51.0 & 114.5 & 130.2 & 13.1 & 123 & 236.7 & 191.3 & -- \\
    \texttt{026\_fan\_chopper\_ota} & 2 & miller & 79.9 & 7.75 & 64.0 & 16.7$^{\P}$ & -1.2$^{\P}$ & 310.1 & 216 & 89.1 & 132.0 & $<$0.001 \\
    \texttt{027\_fan\_rrl\_ota} & 1 & None & 58.7 & 0.15 & 89.8 & 93.1 & 89.6 & 8.2 & 476 & 125.6 & 65.4 & 0.203 \\
    \texttt{031\_srmc\_core\_cmfb} & 2 & miller-rz & 87.6 & 33.20 & 81.3 & $>$180 & $>$180 & 205.0 & 141 & 44.3 & 91.8 & -- \\
    \texttt{032\_ti\_ldo\_error\_selfbias} & 2 & Rz+Cc$^{\ast}$ & 77.0 & 9.75 & 89.8 & 62.4 & 56.4 & 110.7 & 1455 & 54.6 & 103.3 & 0.014 \\
    \texttt{033\_ti\_ldo\_ref\_selfbias} & 2 & Rz+Cc$^{\ast}$ & 78.5 & 6.40 & 72.7 & 44.9 & 46.5 & 73.3 & 343 & 53.4 & 79.3 & 0.242 \\
    \bottomrule
  \end{tabular}}
  \vspace{2pt}

  {\footnotesize \textemdash{} and -- mark a cell with no value: not
  applicable, not recorded in this configuration, or a recorded bench
  failure. $^{\dagger}$AnalogGym-sourced. \(^{\S}\)Re-sized after failing cross-kit verification. The imported \texttt{002\_alfio\_raffc} sizing
  limit-cycled through its RAFFC inner loop and was re-sized; it meets
  every specification except its slew-limited 1.86\,$\mu$s settling
  time. \texttt{017}/\texttt{018} are re-sized inside the kit's
  device-model bins and reported as measured.\par}

  {\footnotesize $^{\ast}$Abbreviated compensation scheme: Rz+Cc = ``Rz + Cc from the first-stage node to the output''.\par}

  {\footnotesize Rejection readings at or above 180\,dB sit at the AC solver's resolution floor and are printed as bounds; every CMRR/PSRR figure is systematic (mismatch excluded), an upper bound (\S\ref{sec:datasheet}).\par}

  {\footnotesize Fully-differential rows report the systematic
  CMRR/PSRR (mismatch excluded), measured on the common-mode path
  with the differential gain folded in.\par}

  {\footnotesize $^{\P}$Chopper core characterized with its output chopper held static; chopping leaves the CM path unchanged, and with no CMFB the rejection is CMFB-limited. Adding one (\texttt{034\_fan\_chopper\_cmfb}) improves both CM-path columns by 43.4\,dB on ihp-sg13g2 for 0.8\,dB of differential gain.\par}

  {\footnotesize PM is the stb loop-probe reading (\texttt{pm\_loop\_deg}) where that bench's unity crossing agrees with the AC sweep to within 20\,\%, and the AC-sweep reading (\texttt{pm\_deg}) otherwise; either is folded into $(-180^\circ,180^\circ]$. Here every tabulated PM cell is the stb loop-probe reading. Both readings are carried per design point in the released scoreboard.\par}

  {\footnotesize Both noise columns are $\mu$V$_{rms}$; THD is \%
  (single-tone, 0.1\,V, or 0.05\,V where the swing would clip, at a
  per-cell frequency set from its UGF, harmonics 2--5).\par}

  {\footnotesize Gate area is the summed gate area of the sized devices only; capacitor and resistor area is excluded (\S\ref{sec:scoreboard}).\par}

  {\footnotesize \texttt{017} is tabulated for information but excluded from the counted corpus; the counted \texttt{028} (behavioral) and \texttt{029} (no recorded baseline) are not tabulated.\par}

  {\footnotesize The $\sim$10\,mW of \texttt{019\_ti\_ldo\_error} is design intent: the source design's 1\,mA reference is mirrored at near-unity ratio into the tail and output legs.\par}

  {\footnotesize $v_{n,\mathrm{in}}$ integrates the uniform 1\,Hz--1\,MHz
  comparison band, $v_{n,\mathrm{in}}^{\mathrm{BW}}$ each cell's own
  1\,Hz--UGF band; the uniform band overstates a slow cell (up to
  7.8$\times$) and understates a fast one (down to 0.17$\times$).\par}
\end{table}

%% file: data/generated/ldo_ppa.tex
\begin{table}[p]
  \caption{LDO benchmark across the three open PDKs at the released
  sizing of each entry (tt corner; top: \texttt{ihp-sg13g2}, middle:
  \texttt{gf180mcu}, bottom: \texttt{sky130}). All values are
  re-simulated, and every entry passes its full datasheet against the
  per-entry bands and operating conditions of Table~\ref{tab:ldo-specs}.
  }\label{tab:ldo-ppa}\label{tab:ldo-ppa-gf}\label{tab:ldo-ppa-sky}
  \centering
  \footnotesize
  \setlength{\tabcolsep}{4pt}
  \textbf{ihp-sg13g2}\par\vspace{2pt}
  \resizebox{\linewidth}{!}{%
  \begin{tabular}{l r r r r r r r r r}
    \toprule
    Circuit & $V_{\mathrm{out}}$ (V) & Dropout (mV) & Load reg.\ (mV) & Line reg.\ (mV) & PSRR (dB) & PM ($^\circ$) & Undershoot (mV) & $I_q$ ($\mu$A) & Gate area ($\mu$m$^2$) \\
    \midrule
    \texttt{001\_analoggym\_basic}$^{\dagger}$\(^{\P}\) & 1.60 & 48 & 26.3 & 4.0 & 37.0 & 89.5 & 161 & 89.1 & 4837 \\
    \texttt{002\_analoggym\_folded\_cascode}$^{\dagger}$\(^{\P}\) & 1.81 & 52 & 11.6 & 0.8 & 48.4 & 97.7 & 632 & 408.1 & 2181 \\
    \texttt{003\_analoggym\_simple}$^{\dagger}$\(^{\P}\) & 1.81 & 148 & 12.5 & 2.9 & 35.2 & 136.0 & 301 & 286.1 & 1758 \\
    \texttt{004\_basic\_pmos}\(^{\P}\) & 1.31 & 41 & 25.1 & 9.5 & 37.4 & 89.3 & 9 & 206.5 & 790 \\
    \texttt{005\_buffered\_ref}\(^{\P}\) & 1.61 & 170 & 1.3 & 4.3 & 44.5 & 46.4 & 2 & 758.7 & 779 \\
    \texttt{007\_pmos}\(^{\P}\) & 1.22 & 99 & 15.7 & 4.7 & 43.9 & 87.0 & 4 & 232.4 & 608 \\
    \texttt{008\_fer\_mirror\_ota} & 1.21 & 76 & 28.0 & 2.0 & 49.5 & 53.7 & 41 & 164.0 & 162 \\
    \texttt{009\_fer\_5t\_pass}\(^{\P}\) & 0.91 & 189 & 35.6 & 0.4 & 60.9 & 90.3 & 19 & 73.9 & 53 \\
    \bottomrule
  \end{tabular}}\par\vspace{8pt}
  \textbf{gf180mcu}\par\vspace{2pt}
  \resizebox{\linewidth}{!}{%
  \begin{tabular}{l r r r r r r r r r}
    \toprule
    Circuit & $V_{\mathrm{out}}$ (V) & Dropout (mV) & Load reg.\ (mV) & Line reg.\ (mV) & PSRR (dB) & PM ($^\circ$) & Undershoot (mV) & $I_q$ ($\mu$A) & Gate area ($\mu$m$^2$) \\
    \midrule
    \texttt{001\_analoggym\_basic}$^{\dagger}$\(^{\P}\) & 1.60 & 51 & 15.8 & 1.4 & 46.6 & 91.4 & 237 & 84.2 & 6663 \\
    \texttt{002\_analoggym\_folded\_cascode}$^{\dagger}$\(^{\P}\) & 1.80 & 101 & 4.8 & 0.6 & 47.9 & 94.1 & 724 & 385.4 & 1460 \\
    \texttt{003\_analoggym\_simple}$^{\dagger}$ & 1.84 & 169 & 59.1 & 6.3 & 29.2 & 89.7 & 994 & 114.2 & 3748 \\
    \texttt{004\_basic\_pmos}\(^{\P}\) & 1.31 & 48 & 11.9 & 0.2 & 71.7 & 87.7 & 6 & 206.5 & 396 \\
    \texttt{005\_buffered\_ref}\(^{\P}\) & 1.64 & 340 & 6.0 & 17.2 & 35.6 & 52.6 & 5 & 472.4 & 900 \\
    \texttt{007\_pmos}\(^{\P}\) & 1.20 & 266 & 5.4 & 0.6 & 69.1 & 49.7 & 7 & 145.6 & 1489 \\
    \texttt{008\_fer\_mirror\_ota} & 1.21 & 99 & 39.0 & 2.5 & 37.4 & 35.3 & 56 & 103.9 & 201 \\
    \texttt{009\_fer\_5t\_pass}\(^{\P}\) & 0.91 & 179 & 24.3 & 0.5 & 62.0 & 87.5 & 13 & 64.9 & 263 \\
    \bottomrule
  \end{tabular}}\par\vspace{8pt}
  \textbf{sky130}\par\vspace{2pt}
  \resizebox{\linewidth}{!}{%
  \begin{tabular}{l r r r r r r r r r}
    \toprule
    Circuit & $V_{\mathrm{out}}$ (V) & Dropout (mV) & Load reg.\ (mV) & Line reg.\ (mV) & PSRR (dB) & PM ($^\circ$) & Undershoot (mV) & $I_q$ ($\mu$A) & Gate area ($\mu$m$^2$) \\
    \midrule
    \texttt{001\_analoggym\_basic}$^{\dagger}$ & 1.60 & 258 & 692.5 & 241.0 & 8.9 & 91.8 & 1243 & 19.0 & 4463 \\
    \texttt{002\_analoggym\_folded\_cascode}$^{\dagger}$ & 1.80 & 82 & 3.6 & 0.3 & 53.2 & 98.1 & 732 & 268.9 & 1425 \\
    \texttt{003\_analoggym\_simple}$^{\dagger}$ & 1.87 & 174 & 100.8 & 10.8 & 25.4 & 90.5 & 920 & 108.4 & 3738 \\
    \texttt{004\_basic\_pmos}\(^{\P}\) & 1.30 & 58 & 12.8 & 0.8 & 64.2 & 86.6 & 6 & 206.5 & 549 \\
    \texttt{007\_pmos}\(^{\P}\) & 1.21 & 238 & 10.7 & 0.5 & 70.1 & 65.1 & 9 & 86.3 & 1260 \\
    \texttt{008\_fer\_mirror\_ota}\(^{\P}\) & 1.21 & 63 & 17.4 & 1.1 & 37.1 & 47.8 & 59 & 101.4 & 162 \\
    \texttt{009\_fer\_5t\_pass}\(^{\P}\) & 0.92 & 355 & 26.1 & 0.3 & 63.4 & 88.3 & 12 & 64.5 & 117 \\
    \bottomrule
  \end{tabular}}\par\vspace{8pt}

{\footnotesize $^{\dagger}$AnalogGym-sourced. \(\P\)Re-sized by the
  per-role \(g_m/I_D\) loop. \texttt{005} carries no sky130 binding and
  the stub \texttt{ldo\_006} is excluded; \texttt{007} sits outside the
  counted corpus of Table~\ref{tab:coverage} but its three bindings are
  tabulated, giving the twenty-three rows. As a capacitor-less regulator
  switching 5--55\,mA into 50\,pF, \texttt{001} has an intrinsically
  large undershoot, inside its deliberately relaxed bands.\par}

  {\footnotesize Judged against one class-level reference band
  ($V_{\mathrm{out}}$ within $\pm5$\,\% of nominal; load and line
  regulation each $\leq 5$\,\% of $V_{\mathrm{out}}$; PSRR $\geq 20$\,dB
  at 1\,kHz; dropout $\leq 300$\,mV; loop phase margin
  $\geq 45^\circ$; $I_q \leq 10$\,\% of nominal load), 10 of the 23
  bindings are within the band: the failures concentrate in the
  $I_q$ (10 of 23) and load regulation (2 of 23) sub-bands, while
  19 of 23 meet the $V_{\mathrm{out}}$/regulation/PSRR/dropout
  core. Nominal here is each entry's own datasheet target rather than
  its measured value: nominal $V_{\mathrm{out}}$ is
  \texttt{metrics.v\_out.spec.typ} and nominal load is
  \texttt{default\_conditions.iload.typical}, which is 5\,mA for
  \texttt{001}, 10\,mA for \texttt{002}/\texttt{003} and 1\,mA for the
  rest, so the $I_q$ sub-band is a different absolute current on different
  rows. Every count here is computed from the released
  scoreboard.\par}
\end{table}